\documentclass[10pt,twocolumn,letterpaper]{article}

\usepackage[pagenumbers]{cvpr} 
\usepackage{graphicx}
\usepackage{capt-of}   
\usepackage{cuted}     

\usepackage[dvipsnames]{xcolor}

\definecolor{cvprblue}{rgb}{0.21,0.49,0.74}
\usepackage[pagebackref,breaklinks,colorlinks,citecolor=cvprblue]{hyperref}

\def\paperID{172} 
\def\confName{3DV\xspace}
\def\confYear{2026\xspace}

\title{Appreciate the View: A Task-Aware Evaluation Framework for Novel View Synthesis}

\author{
Saar Stern$^{1}$  \quad Ido Sobol$^{1}$ \quad Or Litany$^{1,2}$\\
{\small $^{1}$ Technion \quad $^{2}$ NVIDIA
}
}
\newcommand{\datasetname}{\textsc{ViewMatch}\xspace}

\begin{document}
\maketitle
\setlength\stripsep{0pt}
\begin{strip}
    \centering
    \includegraphics[width=0.84\textwidth]{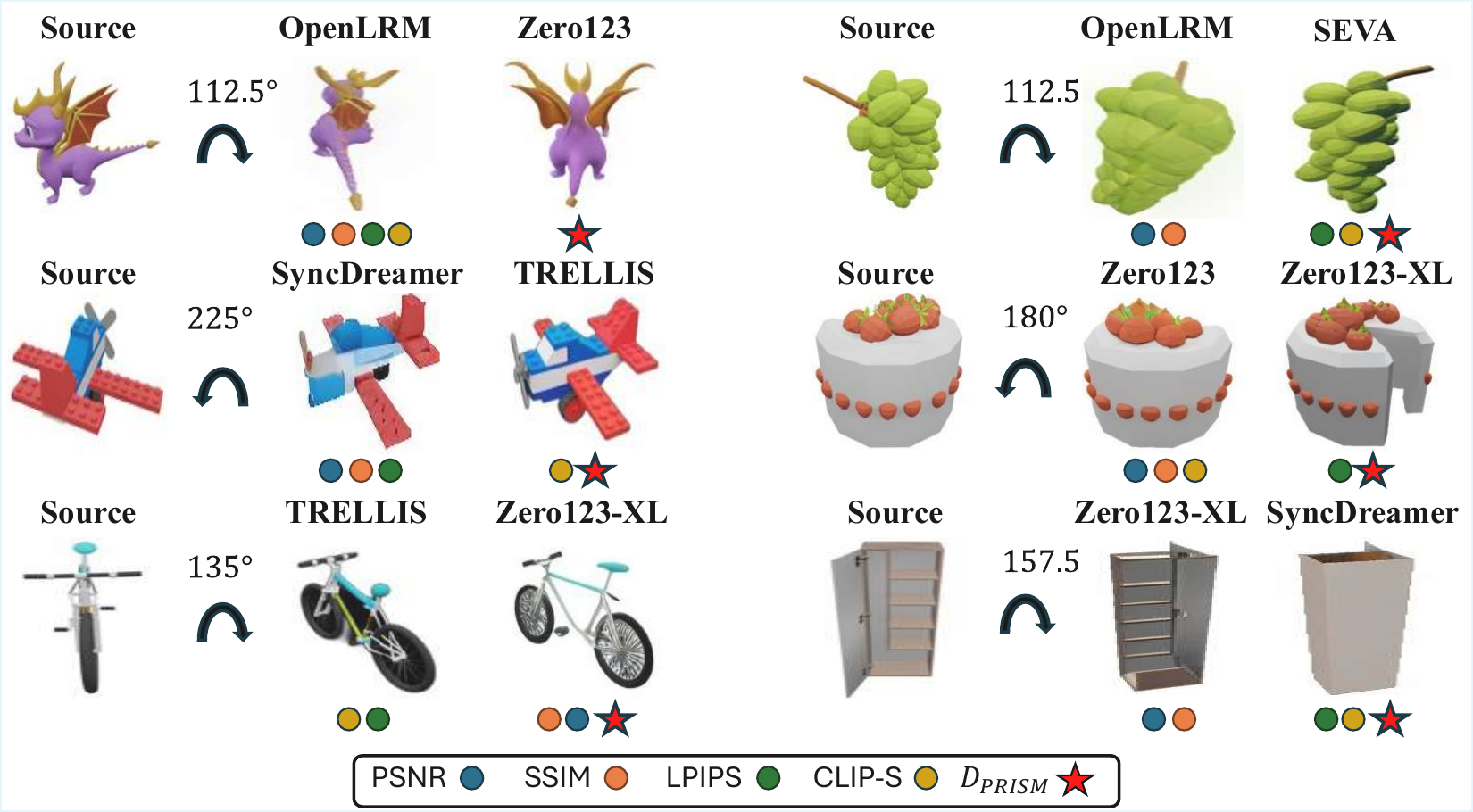}
    \captionof{figure}{Standard metrics (PSNR, SSIM, LPIPS, CLIP-S) often mis-rank incorrect generations in novel view synthesis. 
Our metric, $D_{\text{PRISM}}$, penalizes these incorrect outputs, aligning more closely with human judgments. 
Each pair shows outputs from different NVS models under the same input, with the output favored by each metric indicated.}
\label{fig:teaser}
\end{strip}
\begin{abstract}
The goal of Novel View Synthesis (NVS) is to generate realistic images of a given content from unseen viewpoints. But how can we trust that a generated image truly reflects the intended transformation? Evaluating its reliability remains a major challenge. While recent generative models, particularly diffusion-based approaches, have significantly improved NVS quality, existing evaluation metrics struggle to assess whether a generated image is both realistic and faithful to the source view and intended viewpoint transformation. Standard metrics, such as pixel-wise similarity and distribution-based measures, often mis-rank incorrect results as they fail to capture the nuanced relationship between the source image, viewpoint change, and generated output. 
We propose a task-aware evaluation framework that leverages features from a strong NVS foundation model, Zero123, combined with a lightweight tuning step to enhance discrimination. Using these features, we introduce two complementary evaluation metrics: a reference-based score, $D_{\text{PRISM}}$, and a reference-free score, $\text{MMD}_{\text{PRISM}}$. Both reliably identify incorrect generations and rank models in agreement with human preference studies, addressing a fundamental gap in NVS evaluation. Our framework provides a principled and practical approach to assessing synthesis quality, paving the way for more reliable progress in novel view synthesis. To further support this goal, we apply our reference-free metric to six NVS methods across three benchmarks: Toys4K, Google Scanned Objects (GSO), and OmniObject3D, where $\text{MMD}_{\text{PRISM}}$ produces a clear and stable ranking, with lower scores consistently indicating stronger models.

\end{abstract}
\noindent\textbf{Project page:} \url{https://saarst.github.io/appreciate-the-view-website/}
    
\section{Introduction}
\label{sec:intro}
Novel View Synthesis (NVS) is a fundamental problem in computer vision, requiring the generation of high-quality images from unseen viewpoints while maintaining structural, visual, and semantic consistency with the source view. This capability is essential for applications such as virtual and augmented reality (AR) \cite{andersen2018ar}, as well as robotics \cite{maggio2023loc}, where reconstructing objects and scenes from sparse observations is critical. Recent advances in generative models, particularly diffusion-based approaches \cite{ho2020denoising, song2020denoising}, have significantly improved NVS, enabling realistic image synthesis from limited inputs and accommodating large viewpoint shifts \cite{watson2022novel,liu2023zero}. However, these models remain imperfect, often struggling with geometric consistency and appearance preservation even under mild viewpoint changes. Several strategies have been proposed to improve view synthesis capabilities including using more training data \cite{deitke2023objaverse}, training multi-view models \cite{zhou2025stable}, incorporate strong 3D priors \cite{xiang2024structured} and training-free approaches that enhance generation at inference time \cite{sobol2024zero}. However, progress is hindered by the lack of robust evaluation --- without reliable metrics, improvements remain ambiguous.
Existing metrics fail to capture the relationship between the source image, the intended viewpoint transformation, and the generated output. Reference-based metrics such as PSNR, SSIM~\cite{wang2004image}, and LPIPS~\cite{zhang2018unreasonable} require a ``true'' target image and incorrectly penalize plausible variations that deviate from it. Reference-free metrics like FID~\cite{heusel2017gans} evaluate the generated view in isolation, ignoring consistency with the source view. Consistency-based metrics, such as Met3r~\cite{asim2025met3r}, rely on correspondences between overlapping regions of the source and target, which makes them unreliable for large viewpoint changes where overlap is minimal or absent.

To address this gap, we advocate for the use of deep features, which have become standard in evaluating image generation and reconstruction quality \cite{heusel2017gans,jayasumana2024rethinking}. Specifically, we seek features that encode the source–target–viewpoint relationship, ensuring sensitivity to degradations in consistency and quality. We hypothesize such features already exist in strong NVS backbones, and leverage Zero123-XL~\cite{liu2023zero}. Notably, diffusion model features have recently been shown to possess strong discriminative power~\cite{tang2023emergent, zhan2025general, xu20243difftection}, making them a promising basis for evaluating NVS quality.

To validate whether these features capture the desired relationships, we construct a new dataset, \datasetname, consisting of source images paired with positive and negative target views. Positive samples preserve visible regions from the source and plausibly inpainted occluded parts, while negative samples contain alterations to visible regions that break consistency. This benchmark enables quantitative assessment of whether features separate valid from invalid generations — a key requirement for reliable evaluation.
Our results show that, unlike conventional features \cite{radford2021learning, oquab2023dinov2} that lack viewpoint awareness, ours separate plausible from implausible views; lightweight tuning on the train split of \datasetname further improves discrimination. This requires minimal training data yet improves discrimination between correct and incorrect samples. 

Using the refined features, we form two task aware evaluation metrics: $\text{D}_{\text{PRISM}}$ and $\text{MMD}_{\text{PRISM}}$ which are reference-based and reference-free, respectively. 

To assess the quality of our refined features, we conducted a human study in which participants compared NVS results across multiple models. Our reference-based metric produces rankings of NVS models that align well with human preferences.
Our main contributions are:

\begin{itemize}[nosep,leftmargin=*]

\item \textbf{Leveraging features from an NVS foundation model} to assess adherence to both the source image and the relative viewpoint transformation, providing an evaluation signal suitable for a generative task.
\item \textbf{Constructing a benchmark} of carefully generated positive and negative pairs to (a) systematically tests whether candidate evaluation methods can distinguish between plausible and implausible novel views; (b) fascilitate a lightweight finetuning that enhances the discriminative power of our extracted features.

\item \textbf{Demonstrating the effectiveness of our proposed metric} in ranking NVS models, aligning closely with human preferences.
\end{itemize}

\noindent By addressing a fundamental gap in NVS evaluation, our approach provides a robust metric for assessing synthesis quality, paving the way for more reliable advancements in the field. To further support this goal, we apply our reference-free metric to six NVS methods across three benchmarks: Toys4K, Google Scanned Objects (GSO), and OmniObject3D, where $\text{MMD}_{\text{PRISM}}$ produces a clear and stable ranking, with lower scores consistently indicating stronger models.

\begin{figure*}[h]
    \centering
    \includegraphics[width=0.9\linewidth]{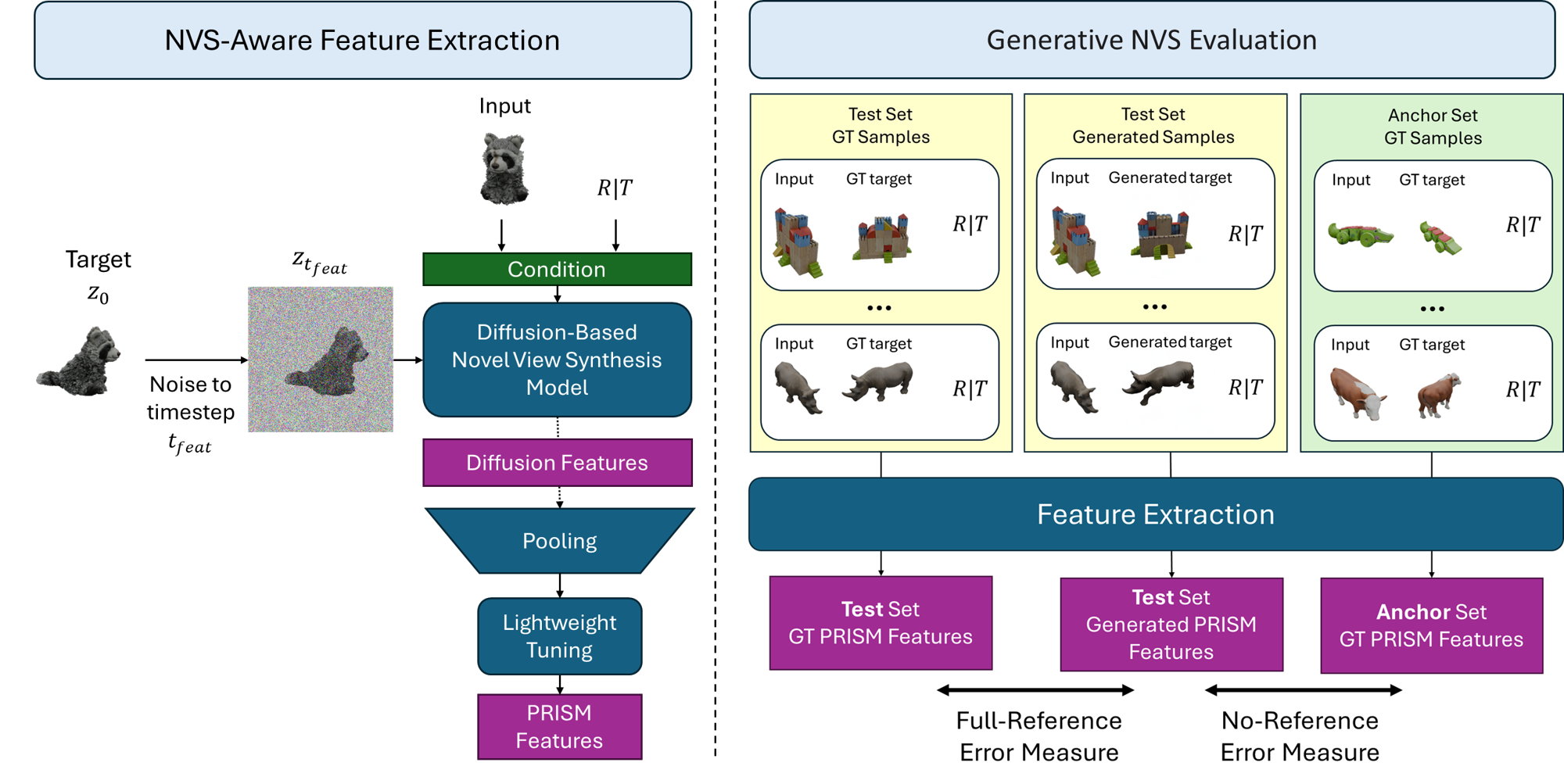}
    \caption{\textbf{Method Overview}. (Left) Feature extraction: given source, target, and camera transformation, we noise the target image and extract features from a diffusion-based NVS model. These are pooled and tuned into \( f_{\text{PRISM}} \). (Right) Evaluation framework: \textbf{Full-Reference:} measure distance between \( f_{\text{PRISM}} \) of a predicted triplet and its ground-truth counterpart. \textbf{No-Reference:} compute MMD between \( f_{\text{PRISM}} \) from generated triplets and an anchor set of real triplets.}
    \label{fig:method}
\end{figure*}

\section{Related Work}
\label{sec:related-work}
\subsection{Generative Novel View Synthesis}
Recent advances in generative modeling \cite{ho2020denoising, rombach2022high} have enabled impressive progress in novel view synthesis (NVS), where the goal is to render unseen views from sparse observations. Methods can be grouped into two categories. Image-based approaches synthesize views directly in pixel (or latent) space, from single-image models such as Zero123 \cite{liu2023zero} to multi-view extensions \cite{liu2023syncdreamer, zhou2025stable}. In contrast, 3D-based methods leverage geometry representations—radiance fields \cite{mildenhall2021nerf}, meshes, Gaussian splats \cite{kerbl20233d}, or latent geometry spaces \cite{xiang2024structured, zhang2024clay}. Other work optimizes 3D representations under 2D supervision via score distillation sampling (SDS) \cite{poole2022dreamfusion, metzer2023latent}. Despite differences in representation and training, all aim to generate plausible novel views, and thus fall within the scope of our evaluation framework.

\subsection{Evaluating Generative Image Models}
Evaluating NVS remains challenging because multiple outputs can be correct.  

\noindent\textbf{Full-reference metrics} such as PSNR, SSIM \cite{wang2004image}, LPIPS \cite{zhang2018unreasonable} and CLIP-S~\cite{hessel2021clipscore} compare against ground-truth targets, but penalize valid variations. Masked or silhouette-based extensions reduce bias \cite{kerbl20233d, sobol2024zero}, yet remain limited. Multi-view consistency methods \cite{woo2024harmonyview, chan2023generative} require multiple target views, making them impractical for single-view evaluation.  

\noindent\textbf{No-reference metrics} avoid the need for ground truth, typically comparing feature distributions (IS, FID, KID, CMMD \cite{salimans2016improved, heusel2017gans, binkowski2018demystifying, jayasumana2024rethinking}). However, they ignore the conditioning source and viewpoint transformation, rewarding realism over correctness. JFID and JFDD \cite{elata2024novel} incorporate source–target pairs but does not explicitly account for viewpoint change. Geometry-aware approaches such as 3DiM \cite{watson2022novel}, TSED \cite{yu2023long}, and Met3r \cite{asim2025met3r} enforce multi-view consistency but either require many views or overlook the plausibility of hallucinated regions. Overall, existing metrics fail to provide reliable single-source-to-single-target evaluation.

\subsection{Diffusion Features for Discriminative Tasks}
Diffusion features capture rich structural and semantic cues and have proven useful for tasks such as correspondence, segmentation, and shape matching \cite{tang2023emergent, xu20243difftection}. Recent work shows that pooled diffusion features support 3D reasoning and lightweight adaptation \cite{zhan2025general, luo2023diffusion}. Applied to NVS, these features naturally encode the interplay between source, target, and viewpoint, making them a promising foundation for evaluation.

\section{Novel View Synthesis Diffusion Features}
\label{sec:nvs-diffusion-features}

In this section, we introduce \textbf{PRISM} (\textit{Pose-aware Representation for Image Synthesis Monitoring})—a compact, triplet-aware embedding for evaluating single-source to single-target novel view synthesis (NVS). Given a source image $I_{\text{src}}$, a target image $I_{\text{tgt}}$, and a known relative transformation $\pi$, NVS aims to generate $I_{\text{tgt}}$ consistent with both $I_{\text{src}}$ and $\pi$. Since the task is ill-posed, evaluation must account for both geometric alignment and the plausibility of synthesized, unobserved regions. PRISM encodes this triplet into a fixed-length feature $f_{\text{PRISM}}(I_{\text{src}}, I_{\text{tgt}}, \pi)$ that enables robust, discriminative assessment.

\begin{figure*}[ht]
    \centering
    \includegraphics[width=\linewidth]{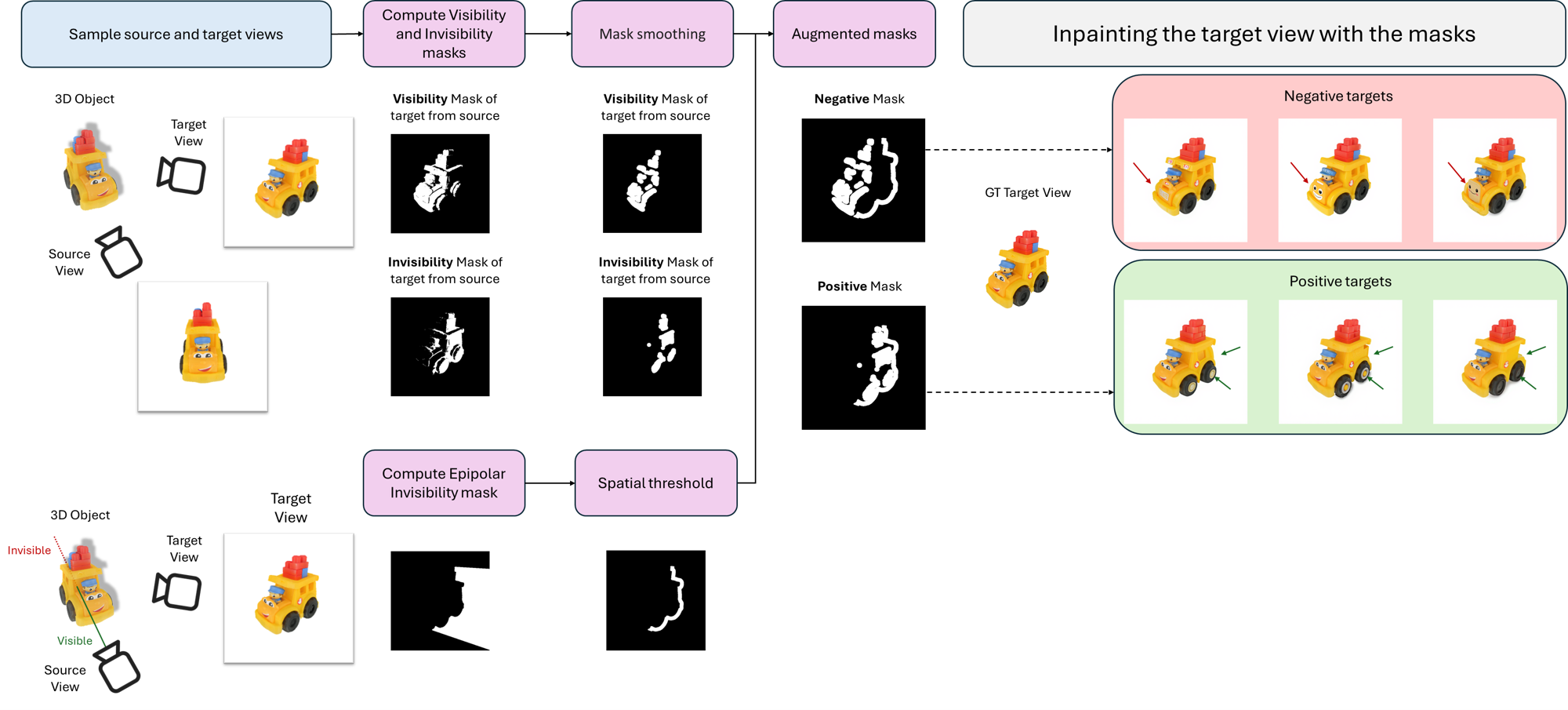}
    \caption{Overview of our \datasetname creation process of positive and negative target examples. (Top Left) Given a 3D mesh and source and target viewpoints, we extract visibility and invisibility masks of the target view from the source, based on the visible faces of the target from the source. (Bottom Left) Given a 3D mesh and source and target viewpoints, we extract an epipolar invisibility mask, representing the unseen regions from the target view, beyond the object. (Right) We augment the visibility and invisibility masks with parts of the epipolar masks, to enable shape changes, and pass the true target and the created masks to an inpainting model.}
    \label{fig:data_creation}
\end{figure*}

Recent work has shown that diffusion models contain rich geometric signals, often accessed by extracting intermediate features and training task-specific classifiers~\cite{zhan2025general}. We extend this idea to the novel view synthesis setting and instead learn a compact embedding via contrastive training, enabling representation-level reasoning over source, pose, and target image triplets.

An overview of the PRISM feature extraction pipeline is shown on the left side of~\cref{fig:method}. At a high level, PRISM uses a conditional NVS diffusion model to extract multi-scale features from an NVS triplet and projects them into a low-dimension $l_2$-normalized embedding space. This projection is trained using a curated dataset of positive and negative triplets (see~\cref{sec:dataset}), where positive triplets correspond to plausible synthesized views, and negatives to implausible ones -- generated under the same input conditions but containing geometric or appearance distortions. This enables the embedding to capture fine-grained differences in task adherence and forms the basis for our evaluation pipeline. Once extracted, PRISM can be applied in two evaluation modes as illustrated on the right side of~\cref{fig:method}:

\noindent\textbf{Reference-Based Evaluation.} Given a predicted triplet $(I_{\text{src}}, \tilde{I}_{\text{tgt}}, \pi)$ and a corresponding ground-truth triplet $(I_{\text{src}}, I_{\text{tgt}}, \pi)$, we compute the  distance between their PRISM embeddings, normalized to $[0,1]$:

\begin{equation}
    \text{D}_{\text{PRISM}} = \frac{1}{2} \left\| f_{\text{PRISM}}(I_{\text{src}}, \tilde{I}_{\text{tgt}}, \pi) -  f_{\text{PRISM}}(I_{\text{src}}, I_{\text{tgt}}, \pi) \right\|
\end{equation}
where lower values indicate stronger alignment with the reference.

\noindent\textbf{Reference-Free Evaluation.}  We compare sets of PRISM embeddings from generated triplets $\mathcal{F}_{\text{gen}}$ and an anchor set $\mathcal{F}_{\text{anch}}$, using maximum mean discrepancy (MMD):
\begin{equation}
    \text{MMD}_{\text{PRISM}} = \text{MMD}(\mathcal{F}_{\text{gen}}, \mathcal{F}_{\text{anch}})
\end{equation}
Lower values indicate stronger alignment. MMD estimates a distance between the underlying distributions, We choose it over Frechet distance (FD) due to its efficient GPU implementation, following CMMD~\cite{jayasumana2024rethinking}; full details are provided in the Appendix~\cref{app:fd_mmd}.

The next two subsections describe the construction of PRISM in detail. In~\cref{subsec:diffusion_features_extraction}, we explain how intermediate triplet-aware features are extracted from a conditional diffusion model. In~\cref{subsec:contrastive_finetuning}, we outline the contrastive training procedure used to project these features into the final embedding space.

\subsection{Diffusion Features Extraction}
\label{subsec:diffusion_features_extraction}

We extract features from a conditional diffusion model trained for novel view synthesis. While our framework is generally applicable, we focus here on Zero123~\cite{liu2023zero} as the backbone, as it is a well-established method for object-level NVS trained on a large-scale dataset. The target image \( I_{\text{tgt}} \) is encoded into a latent representation \( z_0 \), and Gaussian noise \( \epsilon \sim \mathcal{N}(0, \mathbf{I}) \) is added at one timestep \( t \in \left[0,T\right] \) to obtain:
\begin{equation}
    z_t = \sqrt{\bar{\alpha}_t} z_0 + \sqrt{1 - \bar{\alpha}_t} \, \epsilon.
\end{equation}
We perform a single denoising step with U-Net \( f_\theta \), conditioned on the source image \( I_{\text{src}} \) and relative pose \( \pi \), and extract activations from each model block $b \in [1,B] $ (details about Zero123 U-Net architecture are in the Appendix~\cref{app:appendix-unet}):
\begin{equation}
    F_{b} = f_{\theta_b}(z_t, t; I_{\text{src}}, \pi) \in \mathbb{R}^{H_b \times W_b \times C_b},
\end{equation}
Each activation map \( F_{b}  \) is then normalized and spatial pooled :
\begin{equation}
    v_{b} = \frac{1}{H_b W_b} \sum_{i,j} \frac{F_{b}[i,j,:]}{\|F_{b}[i,j,:]\|},
\end{equation}
and concatenated across blocks into a global vector:
\begin{equation}
    v = \text{Concat}(v_{t,1}, \dots, v_{B}) \in \mathbb{R}^C, \quad C = \sum_{b=1}^B C_b
    \label{eq:v}
\end{equation}

\subsection{The \datasetname Benchmark}
\label{sec:dataset}

Although diffusion features encode rich geometry-aware signals, they are not optimized to separate correct from incorrect generations. To learn such discriminative representations, we require supervision from contrastive pairs. 

We therefore construct \datasetname, which provides aligned positive and negative triplets for training. Positives preserve consistency with the source and camera motion, while negatives introduce deliberate violations in content, geometry, or appearance.

We select 40 objects from the GSO dataset~\cite{downs2022google}, each rendered from multiple viewpoints. For every source–target pair, we create positive and negative examples by inpainting the target view using visibility-based masks. The resulting triplets are split into train and test sets for contrastive tuning and evaluation. See Appendix for full details.

\begin{figure*}[t]
    \centering
    \includegraphics[width=0.9\linewidth]{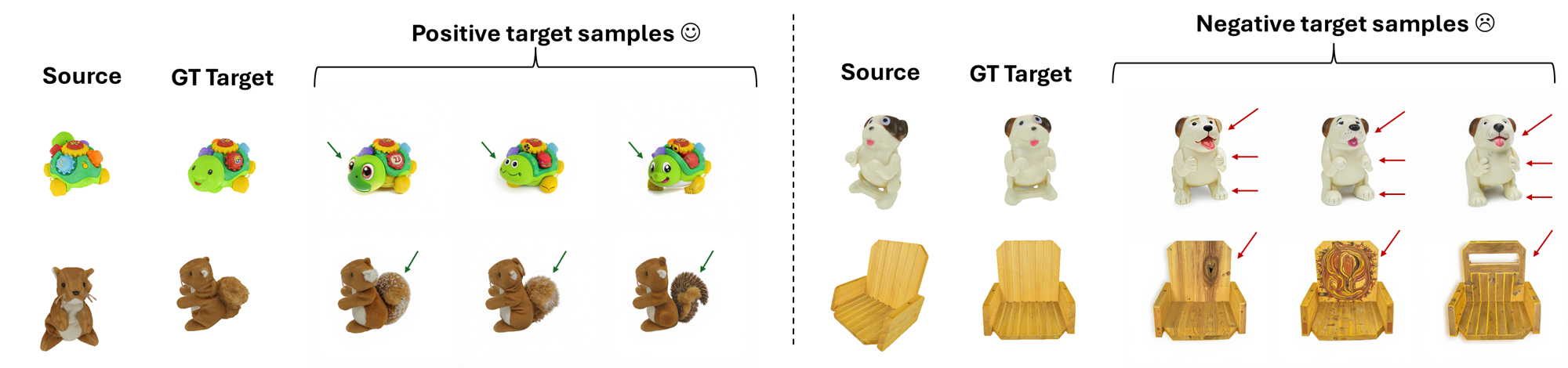}
    \caption{Examples from \datasetname. Each group shows a source view, its ground-truth target, and three generated samples. Positives preserve consistency; negatives violate it through targeted inpainting as described in~\cref{sec:dataset}.}
    \label{fig:dataset}
\end{figure*}

\noindent\textbf{Visibility and Invisibility Masks.} These are masks rendered from the target view, indicating which parts of the object surface are visible from the source view. Specifically, the visibility mask shows which parts in the target view correspond to visible regions from the source view, as these regions should remain unchanged when generating a target view. The invisibility mask shows which parts in the target view correspond to occluded or invisible regions from the source view, as these regions can differ from the specific GT target view while still being geometrically and semantically consistent with the source view and camera transformation. We generate these masks by analyzing per-viewpoint visibility and applying post-processing to refine object boundaries. Further implementation details are provided in the Appendix.

\noindent\textbf{Epipolar Mask.} Inspired by iNVS~\cite{kant2023invs}, this mask is rendered from the target view and highlights regions that lie in 3D volumes invisible from the source view. We cast rays from the source camera onto the visible object surface; each continues into occluded space beyond the first intersection. Segments that pass through the object are discarded, and remaining occluded rays are projected onto the target view. These epipolar masks are used to augment visibility-based masks, enabling more diverse inpainting regions beyond the object silhouette. To avoid unrealistic edits, we constrain these masks to a neighborhood around the object. As these regions are unobserved in the source, any inpainting within them remains geometrically valid.

\noindent\textbf{Generating Positive and Negative Examples.}
We generate positives by inpainting the invisibility and epipolar masks using FLUX~\cite{flux2024}, preserving geometric validity. Negatives are created by inpainting visibility and epipolar masks, corrupting regions that should remain unchanged. In both cases, the inpainting is text-guided using BLIP~\cite{li2022blip} prompts from the source image, ensuring semantic consistency and making the distinction more challenging. The use of epipolar masks extends edits beyond the ground-truth object area, increasing diversity and difficulty. See~\cref{fig:data_creation},~\cref{fig:dataset}, and the appendix for examples.

With these positive and negative triplets available, we can now perform contrastive training, using \datasetname to fine-tune the projection head that maps raw diffusion features into the final PRISM embedding.

\subsection{Contrastive Fine-Tuning}  
\label{subsec:contrastive_finetuning}

The raw diffusion features \(v\) (\cref{eq:v}) encode multi-scale cues from a source--pose--target triplet, but they are not explicitly optimized to distinguish plausible from implausible generations. To specialize them for evaluation, we train a lightweight two-layer MLP with ReLU activation using supervision from \datasetname (Sec.~\ref{sec:dataset}). The MLP projects \(v\) into a compact, \(\ell_2\)-normalized embedding:

\begin{equation}
    f_{\text{PRISM}} = \frac{h(v)}{\| h(v) \|}.
\end{equation}
The compact size of \(f_{\text{PRISM}}\) is important: it allows efficient distributional comparison via MMD, while reducing storage overhead when saving anchor sets.

\noindent Training follows a triplet contrastive strategy. For each GT triplet \((I_{\text{src}}, I_{\text{tgt}}, \pi)\) in the train split of \datasetname, we form contrastive examples by pairing the anchor (GT) with:
\begin{itemize}
    \item \textbf{Positive:} a plausible synthesized target consistent with the source and pose.  
    \item \textbf{Negative:} either (i) an implausible target from \datasetname, or (ii) a target view rendered at the wrong pose ("misaligned-angle").  
\end{itemize}

\noindent The embedding is optimized with a margin-based triplet loss:
\begin{equation}
    \mathcal{L}(a,p,n) = \max\!\big(d(a,p) - d(a,n) + m, \, 0 \big),
\end{equation}
where \(d(s,s') = \|s-s'\|_2\) and \(m\) is a margin hyper-parameter (further details are provided in the experimental setup). This encourages GT-aligned embeddings to be close to positives and far from both \datasetname and misaligned-angle negatives.

\section{Experiments}
\label{sec:experiments}

We evaluate $\text{D}_{\text{PRISM}}$ and $\text{MMD}_{\text{PRISM}}$ across a series of controlled analyses and cross-dataset comparisons.  
The datasets we rely on are:

\begin{itemize}
    \item \textbf{\datasetname.} Our benchmark of positive and inpainted negative triplets, used to test discriminability, assess full-reference metrics, and validate no-reference evaluation.
    \item \textbf{Misaligned-GSO.} A controlled set where ground-truth targets are rendered at incorrect azimuths, isolating viewpoint sensitivity.
    \item \textbf{GSO~\cite{downs2022google}.} Used for ranking (200/200 anchor/test split), for the user study with five diverse NVS models, and for degradation experiments.
    \item \textbf{Toys4K~\cite{Stojanov_2021_CVPR} and OmniObject3D~\cite{wu2023omniobject3d}} Large-scale object datasets; we sample 40 and 20 categories respectively to benchmark rankings across domains.
\end{itemize}

\paragraph{Training Details.}  
We optimize the projection head with AdamW (learning rate $1\!\times\!10^{-4}$, batch size 32) for 100 epochs, using a margin of $m=1$ in the triplet loss.  
The embedding dimension is set to 2048, which is roughly one quarter of the raw diffusion feature size, providing a balance between discriminability and compactness for efficient MMD computation and anchor storage.  
Unless stated otherwise, all reported results use features extracted at $t=0$.

\paragraph{Evaluation Roadmap.}
The subsections that follow build on these datasets in sequence:  
(\cref{subsec:ablation_discriminability}–\cref{subsec:val_viewmatch}) establish discriminability and no-reference validation on \datasetname,  
(\cref{subsec:user-study}) reports alignment with human judgments from the GSO-based study with five NVS models,  
(\cref{subsec:nvs_ranking}) ranks NVS models across GSO, Toys4K, and OmniObject3D,  
and (\cref{sec:additional_analysis}) analyzes robustness to pose misalignment and image degradation.

\subsection{Discriminability of NVS Diffusion Features}
\label{subsec:ablation_discriminability}

We first test whether raw diffusion features can distinguish valid from invalid targets in \datasetname. Baselines use CLIP and DINOv2 embeddings of source and target images concatenated with the relative angle (\textbf{CAT-Angle}). As shown in~\cref{tab:combined_metrics} (right), diffusion features achieve 0.90 AUC, clearly outperforming CLIP (0.73) and DINOv2 (0.68). Results are stable across diffusion timesteps ($t=0,100,200$), with full results in the Appendix. For efficiency, we use $t=0$ features in the remainder of the paper. This demonstrates that even without tuning, diffusion features encode viewpoint-sensitive discriminative signals absent from standard image features.

\begin{table}[t]
\centering
\small
\caption{Comparison of standard metrics and linear classifiers on \datasetname. Left: average scores for pointwise evaluation metrics on positive (P) and negative (N) samples. Right: AUC scores for linear classifiers trained to distinguish positive from negative triplets.}
\label{tab:combined_metrics}
\setlength{\tabcolsep}{6pt}
\begin{tabular}{lcc lc}
\toprule
\multicolumn{3}{c}{\textbf{Pointwise Evaluation Metrics}} & \multicolumn{2}{c}{\textbf{Linear Classifier AUC $\uparrow$}} \\
\cmidrule(r){1-3} \cmidrule(l){4-5}
\textbf{Metric} & \textbf{P} & \textbf{N} & \textbf{Feature} & \textbf{AUC} \\
\midrule
PSNR (dB) $\uparrow$ & 18.18 & 18.69 & CLIP-CAT-Angle  & 0.73 \\
SSIM $\uparrow$      & 0.852 & 0.835 & DINO-CAT-Angle  & 0.68 \\
LPIPS $\downarrow$   & 0.163 & 0.178 & Diffusion ($t=0$)    & 0.90 \\
CLIP-S $\uparrow$    & 0.89  & 0.96  & Diffusion ($t=100$)  & 0.87 \\
$\text{D}_{\text{PRISM}}$ $\downarrow$ & \textbf{0.299} & \textbf{0.703} & Diffusion ($t=200$)  & 0.86 \\
\bottomrule
\end{tabular}
\end{table}

\subsection{Limitations of Standard Full-Reference Metrics}
\label{subsec:baseline_metrics}

A natural question is whether standard full-reference (FR) metrics can identify invalid targets. 
We therefore benchmark PSNR, SSIM, LPIPS, and CLIP-S on positive versus inpainted negative triplets. 
As shown in~\cref{tab:combined_metrics} (left), these metrics not only fail to distinguish the two, but in some cases even assign \textit{higher} scores to negatives. 
This is because both positive and negative samples remain perceptually close to the ground-truth target, so pixel- or feature-based similarity scores provide little separation. 
These findings illustrate that conventional FR metrics are poorly suited to NVS, where validity depends on viewpoint and structural consistency rather than low-level similarity.

\subsection{Reference-Free Distributional Evaluation}
\label{subsec:val_viewmatch}

We next ask whether $ \text{MMD}_{\text{PRISM}} $ can serve as a no-reference (NR) evaluation metric, distinguishing valid from invalid generations without ground-truth targets. To test this, we compare distributions of PRISM features from positive versus negative triplets in \datasetname against an anchor distribution built from 400 held-out objects.

As shown in~\cref{tab:viewmatch_val}, positives achieve substantially lower $ \text{MMD}_{\text{PRISM}} $ scores (0.69) than negatives (0.98), indicating reliable separation at the distributional level. In contrast, standard NR metrics such as FID, CMMD, and JFID yield nearly identical scores for positives and negatives. Because these baselines operate purely on image appearance, they ignore the viewpoint-conditioned structure of NVS and thus fail to penalize implausible generations. This highlights that $ \text{MMD}_{\text{PRISM}} $ provides a practical reference-free evaluation signal sensitive to both plausibility and geometric consistency.

\begin{table}[t]
\centering
\caption{No-reference validation on \datasetname. Lower is better.}
\label{tab:viewmatch_val}
\setlength{\tabcolsep}{5pt}
\small
\begin{tabular}{lcc}
\toprule
\textbf{Method} & \textbf{P} & \textbf{N} \\
\midrule
FID    & 102.360 & 107.240 \\
CMMD   & 0.814   & 0.792   \\
FDD    & 0.701   & 0.725   \\
\midrule
JFID   & 259.120 & 265.180 \\
JCMMD  & 1.041   & 1.024   \\
JFDD   & 1.631   & 1.659   \\
\midrule
$\text{MMD}_{\text{PRISM}}$ & \textbf{0.691} & \textbf{0.984} \\
\bottomrule
\end{tabular}
\end{table}

\subsection{Alignment with Human Judgments}
\label{subsec:user-study}

Human perception is the gold standard for assessing highly generative outputs. We therefore test whether $D_{\text{PRISM}}$ aligns with human preferences. Each task presents a source image, a relative camera motion, and two candidate target views to be judged along four aspects: \textit{Viewpoint Accuracy}, \textit{Shared Region Consistency}, \textit{Plausibility of New Regions}, and \textit{Image Quality}. A blurred hint of the ground-truth target aids viewpoint assessment while concealing fine detail.

We conducted the study on 40 held-out objects from the GSO dataset, distinct from those used in the creation of \datasetname. 
For each object we rendered two large azimuth shifts ($>90^\circ$), yielding 320 possible pairwise comparisons across five diverse NVS models—OpenLRM, Zero123, Zero123-XL, TRELLIS, and SEVA—covering regression-based, multi-view, and 3D generation approaches. 
Forty participants each completed 15 comparison tasks, producing just over 500 pairwise judgments; with four questions per task, this corresponds to roughly 2{,}000 aspect-level human judgments. 
Full details on interface, instructions, and blur generation are provided in Appendix~\cref{app:user_study}.

\cref{tab:pearson_correlation} reports Pearson correlation between metric predictions and human majority votes. 
Standard full-reference metrics (PSNR, SSIM, LPIPS) achieve weak or even negative correlation, while MEt3R and CLIP-S also fail to capture human choices. 
In contrast, $D_{\text{PRISM}}$ ranks first in 3 of the 4 aspects and second in the remaining one. 
Although absolute correlation values remain modest—reflecting the inherent difficulty of evaluating generative tasks—this still represents a clear advance, underscoring the suitability of $D_{\text{PRISM}}$ as a perceptual evaluation metric for NVS.

\begin{table}[t]
\centering
\caption{\textbf{Pearson correlation} between metric predictions and human judgments. 
Columns denote: VP = Viewpoint Accuracy, SC = Shared Consistency, PL = Plausibility of New Regions, IQ = Image Quality. Higher is better.}
\label{tab:pearson_correlation}
\setlength{\tabcolsep}{6pt}
\small
\begin{tabular}{lcccc}
\toprule
\textbf{Metric} & \textbf{VP} & \textbf{SC} & \textbf{PL} & \textbf{IQ} \\
\midrule
PSNR             & 0.071  & 0.007  & -0.187 & -0.323 \\
SSIM             & \textbf{0.279}  & \underline{0.270}  & 0.114  & -0.117 \\
LPIPS            & 0.071  & 0.056  & -0.035 & -0.251 \\
MEt3R            & -0.028 & -0.016 & \underline{0.179}  & 0.080 \\
CLIP-S w/ GT     & -0.330 & 0.107  & -0.084 & \underline{0.019} \\
CLIP-S w/ src    & -0.012 & -0.048 & -0.034 & -0.008 \\
$\text{D}_{\text{PRISM}}$ (Ours) & \underline{0.223} & \textbf{0.352} & \textbf{0.205} & \textbf{0.394} \\
\bottomrule
\end{tabular}
\end{table}

\subsection{Reference-Free Ranking of NVS Models}
\label{subsec:nvs_ranking}

We use $\text{MMD}_{\text{PRISM}}$ to provide a reference-free leaderboard for single-view NVS models. 
Our evaluation covers six methods: Zero123, Zero123-XL, SEVA, OpenLRM, TRELLIS, and SyncDreamer~\cite{liu2023syncdreamer}. 
Rankings are computed on GSO, Toys4K, and OmniObject3D, using the anchor–test splits described in Section~\ref{sec:experiments}. 
For each model we compute $\text{MMD}_{\text{PRISM}}$ between its generated triplets and the dataset-specific anchor distribution, and report \emph{distribution-based baselines} (FID, CMMD, FDD, and their joint variants JFID, JCMMD, JFDD, more details in Appendix~\cref{app:fd_mmd}).

\cref{tab:toys4k} reports results on Toys4K, where $\text{MMD}_{\text{PRISM}}$ produces a clear and consistent ranking, with lower scores corresponding to stronger models. In contrast, the distribution-based baselines often yield compressed score ranges that make it difficult to distinguish performance levels. Importantly, this trend is not unique to Toys4K: additional experiments on GSO and OmniObject3D (\cref{tab:gso_ranking,tab:omniobject3d}), with summary results shown in \cref{tab:all_datasets}, confirm that $\text{MMD}_{\text{PRISM}}$ provides stable rankings across datasets of varying difficulty, suggesting its practicality as a general, reference-free tool for comparing NVS models.

\paragraph{Discussion.} Qualitative inspection (\cref{fig:teaser}) further supports the reported rankings (e.g.~\cref{tab:toys4k}) through highlighting systematic differences across models. OpenLRM underperforms relative to generative methods, consistent with its feed-forward design, which limits its ability to capture diverse object appearances. TRELLIS, though producing visually compelling outputs, is an image-to-3D model repurposed for NVS via source-view angle alignment; small angle mismatches introduced in this process reduce its score under PRISM, which is sensitive to pose inconsistency. Importantly, although PRISM is built on Zero123-XL features, it does not bias toward that model: methods such as SEVA and SyncDreamer consistently achieve higher scores across all three benchmarks. Taken together, these findings demonstrate that PRISM provides a robust and reliable evaluation tool. 

\begin{table}[t]
    \centering
    \caption{\textbf{Toys4K benchmark.} Comparison of reference-free metrics for ranking NVS models. Lower is better.}
    \label{tab:toys4k}
    \small
    \setlength{\tabcolsep}{4pt}
    \resizebox{\linewidth}{!}{
        \begin{tabular}{lcccccc}
        \toprule
        \textbf{Metric} & \textbf{OpenLRM} & \textbf{Z123} & \textbf{Z123-XL} & \textbf{TRELLIS} & \textbf{SEVA} & \textbf{SyncDreamer} \\
        \midrule
        FID    & 168.5738 & 163.3072 & 161.2755 & 164.3074 & 156.2047 & 160.4168 \\
        CMMD   & 118.6056 & 85.6184  & 86.1490  & 34.7009  & 37.7579  & 102.9813 \\
        FDD    & 0.7483   & 0.7389   & 0.7249   & 0.6690   & 0.7030   & 0.7273   \\ 
        \midrule
        JFID   & 379.3934 & 371.8979 & 369.4718 & 373.8946 & 364.4985 & 371.3178 \\
        JCMMD  & 1.0403   & 0.7972   & 0.8036   & 0.3383   & 0.3680   & 0.9351   \\
        JFDD   & 1.5605   & 1.5380   & 1.5188   & 1.4673   & 1.5002   & 1.5350   \\
        \midrule
        $\text{MMD}_{\text{PRISM}}$ (Ours) 
               & 0.8017   & 0.3930   & 0.3067   & 0.4304   & 0.1978 & 0.2216 \\
        \bottomrule
        \end{tabular}
    }
\end{table}

\begin{table}[t]
    \centering
    \small
    \setlength{\tabcolsep}{5pt}
    \renewcommand{\arraystretch}{1.05}
    \caption{$\text{MMD}_{\text{PRISM}}$ results across \textbf{Toys4K}, \textbf{GSO}, and \textbf{OmniObject3D}. 
    For GSO and OmniObject3D, results of alternative metrics are deferred to the appendix.}
    \label{tab:all_datasets}
    \resizebox{\linewidth}{!}{
        \begin{tabular}{lcccccc}
        \toprule
        \textbf{Dataset} & \textbf{OpenLRM} & \textbf{Z123} & \textbf{Z123-XL} & \textbf{TRELLIS} & \textbf{SEVA} & \textbf{SyncDreamer} \\
        \midrule
        Toys4K        & 0.8017 & 0.3930 & 0.3067 & 0.4304 & 0.1978 & 0.2216 \\
        GSO           & 0.8415 & 0.3552 & 0.2997 & 0.2858 & 0.1231 & 0.1392 \\
        OmniObject3D  & 1.3752 & 0.6376 & 0.5206 & 0.5696 & 0.2896 & 0.2030 \\
        \bottomrule
        \end{tabular}
    }
\end{table}

\subsection{Additional Analysis}
\label{sec:additional_analysis}

\begin{figure}[]
    \centering
    \includegraphics[width=0.6\linewidth]{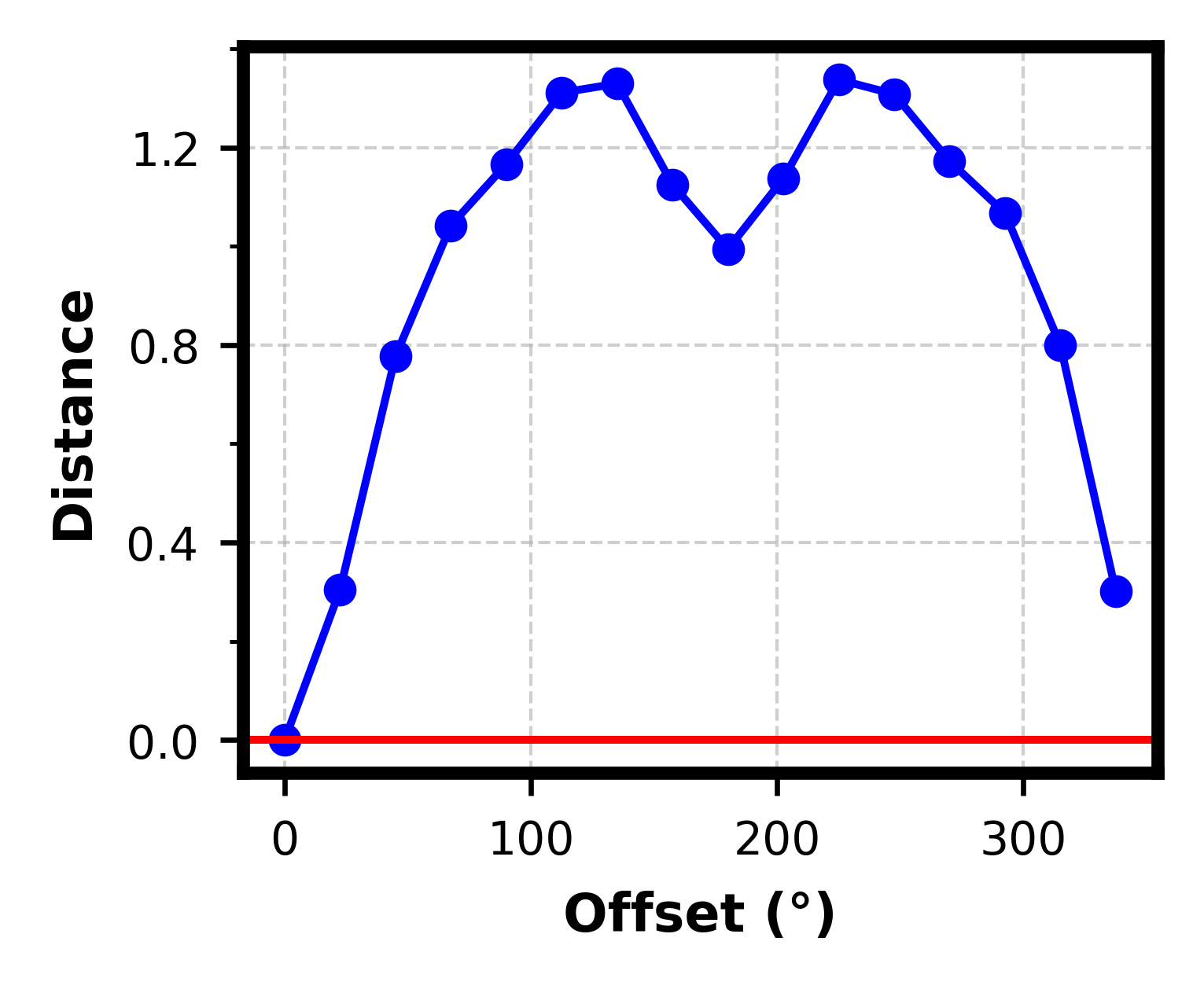}
    \caption{
    Metric scores vs. azimuth offset (0°–360°). 
    Blue: our $\text{MMD}_{\text{PRISM}}$; 
    Red: distribution-based baselines (e.g., FID, JFID). 
    While the baselines remain flat, $\text{MMD}_{\text{PRISM}}$ shows an M-shaped curve, indicating pose sensitivity.
    }
    \label{fig:pose_no_ref}
\end{figure}

\paragraph{Sensitivity to Pose Misalignment.}
We evaluate pose sensitivity on a controlled set of ground-truth renderings, where target views are intentionally misaligned by varying azimuth offsets relative to the source. 
Both $D_{\text{PRISM}}$ and $\text{MMD}_{\text{PRISM}}$ follow an M-shaped trend, rising with larger offsets, peaking near 180°, dipping at symmetries, and recovering toward 360°, mirroring PSNR, SSIM, and LPIPS (Appendix~\cref{fig:pose_full_ref}) and confirming that our features capture viewpoint structure. 
In contrast, no-reference baselines (e.g., FID, JFID) remain flat (\cref{fig:pose_no_ref}), as they ignore source conditioning and thus score even misaligned ground-truth views as plausible. 
This underscores the need for viewpoint-aware evaluation.

\begin{figure}
    \centering
    \includegraphics[width=0.75\linewidth]{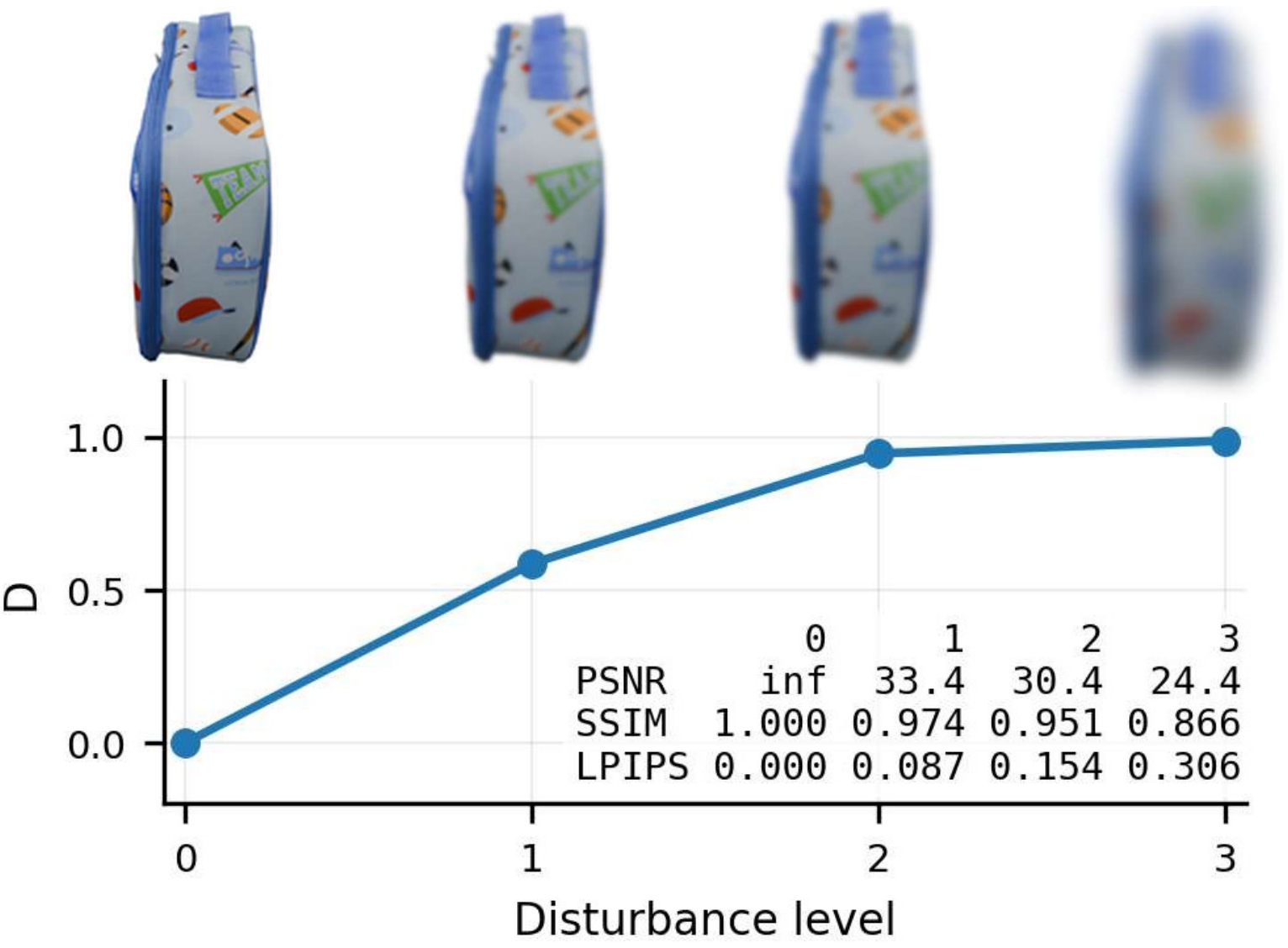}
    \caption{Degradation of $D_{\text{PRISM}}$ (denoted as $D$ in the plot) under Gaussian blur at increasing intensity levels.}
    \label{fig:blur_main}
\end{figure}

\paragraph{Sensitivity to Image Degradation.}
We also test responsiveness to low-level corruptions. 
Following \cite{devries2019evaluation}, we apply Gaussian noise, Gaussian blur, color shifts, and salt-and-pepper noise at increasing intensities (parameters in Appendix~\cref{app:image_degradations}). 
Results for Gaussian blur are shown in \cref{fig:blur_main}, while the other corruptions appear in \cref{fig:color}--\ref{fig:saltpepper}. 
Across all cases, $D_{\text{PRISM}}$ degrades consistently, with scores increasing steadily as distortions intensify, confirming its sensitivity to perceptual quality.

\section{Conclusion and Limitations}
\label{sec:Conclusion}

We introduced task-aware diffusion features for NVS evaluation, explicitly capturing source–target–viewpoint relationships with a strong diffusion backbone and lightweight adaptation. Our metrics, $D_{\text{PRISM}}$ and $\text{MMD}_{\text{PRISM}}$, show higher sensitivity to implausible generations than existing baselines, and enable both reference-based and reference-free evaluation. These tools provide a first step toward more reliable benchmarking of single-view NVS models, which remains a critical need for the community.

A limitation of our approach is its reliance on the Zero123-XL backbone, though the framework can be readily extended to other NVS models. Future work includes applying it to scene-level and multi-view settings.

\section{Acknowledgement}
Or Litany acknowledges support from the Israel Science Foundation (grant 624/25) and the Azrieli Foundation Early Career Faculty Fellowship. This research was also supported in part by an academic gift from Meta. The authors gratefully acknowledge this support.
{
    \small
    \bibliographystyle{ieeenat_fullname}
    \bibliography{main}

\begin{thebibliography}{46}
\providecommand{\natexlab}[1]{#1}
\providecommand{\url}[1]{\texttt{#1}}
\expandafter\ifx\csname urlstyle\endcsname\relax
  \providecommand{\doi}[1]{doi: #1}\else
  \providecommand{\doi}{doi: \begingroup \urlstyle{rm}\Url}\fi

\bibitem[Andersen and Popescu(2018)]{andersen2018ar}
Daniel Andersen and Voicu Popescu.
\newblock An ar-guided system for fast image-based modeling of indoor scenes.
\newblock In \emph{2018 IEEE Conference on Virtual Reality and 3D User Interfaces (VR)}, pages 501--502. IEEE, 2018.

\bibitem[Asim et~al.(2025)Asim, Wewer, Wimmer, Schiele, and Lenssen]{asim2025met3r}
Mohammad Asim, Christopher Wewer, Thomas Wimmer, Bernt Schiele, and Jan~Eric Lenssen.
\newblock Met3r: Measuring multi-view consistency in generated images.
\newblock \emph{arXiv preprint arXiv:2501.06336}, 2025.

\bibitem[Bi{\'n}kowski et~al.(2018)Bi{\'n}kowski, Sutherland, Arbel, and Gretton]{binkowski2018demystifying}
Miko{\l}aj Bi{\'n}kowski, Danica~J Sutherland, Michael Arbel, and Arthur Gretton.
\newblock Demystifying mmd gans.
\newblock \emph{arXiv preprint arXiv:1801.01401}, 2018.

\bibitem[Chan et~al.(2023)Chan, Nagano, Chan, Bergman, Park, Levy, Aittala, De~Mello, Karras, and Wetzstein]{chan2023generative}
Eric~R Chan, Koki Nagano, Matthew~A Chan, Alexander~W Bergman, Jeong~Joon Park, Axel Levy, Miika Aittala, Shalini De~Mello, Tero Karras, and Gordon Wetzstein.
\newblock Generative novel view synthesis with 3d-aware diffusion models.
\newblock In \emph{Proceedings of the IEEE/CVF International Conference on Computer Vision}, pages 4217--4229, 2023.

\bibitem[Deitke et~al.(2023)Deitke, Schwenk, Salvador, Weihs, Michel, VanderBilt, Schmidt, Ehsani, Kembhavi, and Farhadi]{deitke2023objaverse}
Matt Deitke, Dustin Schwenk, Jordi Salvador, Luca Weihs, Oscar Michel, Eli VanderBilt, Ludwig Schmidt, Kiana Ehsani, Aniruddha Kembhavi, and Ali Farhadi.
\newblock Objaverse: A universe of annotated 3d objects.
\newblock In \emph{Proceedings of the IEEE/CVF Conference on Computer Vision and Pattern Recognition}, pages 13142--13153, 2023.

\bibitem[DeVries et~al.(2019)DeVries, Romero, Pineda, Taylor, and Drozdzal]{devries2019evaluation}
Terrance DeVries, Adriana Romero, Luis Pineda, Graham~W Taylor, and Michal Drozdzal.
\newblock On the evaluation of conditional gans.
\newblock \emph{arXiv preprint arXiv:1907.08175}, 2019.

\bibitem[Downs et~al.(2022)Downs, Francis, Koenig, Kinman, Hickman, Reymann, McHugh, and Vanhoucke]{downs2022google}
Laura Downs, Anthony Francis, Nate Koenig, Brandon Kinman, Ryan Hickman, Krista Reymann, Thomas~B McHugh, and Vincent Vanhoucke.
\newblock Google scanned objects: A high-quality dataset of 3d scanned household items.
\newblock In \emph{2022 International Conference on Robotics and Automation (ICRA)}, pages 2553--2560. IEEE, 2022.

\bibitem[Elata et~al.(2024)Elata, Kawar, Ostrovsky-Berman, Farber, and Sokolovsky]{elata2024novel}
Noam Elata, Bahjat Kawar, Yaron Ostrovsky-Berman, Miriam Farber, and Ron Sokolovsky.
\newblock Novel view synthesis with pixel-space diffusion models.
\newblock \emph{arXiv preprint arXiv:2411.07765}, 2024.

\bibitem[He and Wang(2023)]{openlrm}
Zexin He and Tengfei Wang.
\newblock Openlrm: Open-source large reconstruction models.
\newblock \url{https://github.com/3DTopia/OpenLRM}, 2023.

\bibitem[Hessel et~al.(2021)Hessel, Holtzman, Forbes, Bras, and Choi]{hessel2021clipscore}
Jack Hessel, Ari Holtzman, Maxwell Forbes, Ronan~Le Bras, and Yejin Choi.
\newblock Clipscore: A reference-free evaluation metric for image captioning.
\newblock \emph{arXiv preprint arXiv:2104.08718}, 2021.

\bibitem[Heusel et~al.(2017)Heusel, Ramsauer, Unterthiner, Nessler, and Hochreiter]{heusel2017gans}
Martin Heusel, Hubert Ramsauer, Thomas Unterthiner, Bernhard Nessler, and Sepp Hochreiter.
\newblock Gans trained by a two time-scale update rule converge to a local nash equilibrium.
\newblock \emph{Advances in neural information processing systems}, 30, 2017.

\bibitem[Ho et~al.(2020)Ho, Jain, and Abbeel]{ho2020denoising}
Jonathan Ho, Ajay Jain, and Pieter Abbeel.
\newblock Denoising diffusion probabilistic models.
\newblock \emph{Advances in neural information processing systems}, 33:\penalty0 6840--6851, 2020.

\bibitem[Hong et~al.(2023)Hong, Zhang, Gu, Bi, Zhou, Liu, Liu, Sunkavalli, Bui, and Tan]{hong2023lrm}
Yicong Hong, Kai Zhang, Jiuxiang Gu, Sai Bi, Yang Zhou, Difan Liu, Feng Liu, Kalyan Sunkavalli, Trung Bui, and Hao Tan.
\newblock Lrm: Large reconstruction model for single image to 3d.
\newblock \emph{arXiv preprint arXiv:2311.04400}, 2023.

\bibitem[Jayasumana et~al.(2024)Jayasumana, Ramalingam, Veit, Glasner, Chakrabarti, and Kumar]{jayasumana2024rethinking}
Sadeep Jayasumana, Srikumar Ramalingam, Andreas Veit, Daniel Glasner, Ayan Chakrabarti, and Sanjiv Kumar.
\newblock Rethinking fid: Towards a better evaluation metric for image generation.
\newblock In \emph{Proceedings of the IEEE/CVF Conference on Computer Vision and Pattern Recognition}, pages 9307--9315, 2024.

\bibitem[Kant et~al.(2023)Kant, Siarohin, Vasilkovsky, Guler, Ren, Tulyakov, and Gilitschenski]{kant2023invs}
Yash Kant, Aliaksandr Siarohin, Michael Vasilkovsky, Riza~Alp Guler, Jian Ren, Sergey Tulyakov, and Igor Gilitschenski.
\newblock invs: Repurposing diffusion inpainters for novel view synthesis.
\newblock In \emph{SIGGRAPH Asia 2023 Conference Papers}, pages 1--12, 2023.

\bibitem[Kerbl et~al.(2023)Kerbl, Kopanas, Leimk{\"u}hler, and Drettakis]{kerbl20233d}
Bernhard Kerbl, Georgios Kopanas, Thomas Leimk{\"u}hler, and George Drettakis.
\newblock 3d gaussian splatting for real-time radiance field rendering.
\newblock \emph{ACM Trans. Graph.}, 42\penalty0 (4):\penalty0 139--1, 2023.

\bibitem[Labs(2024)]{flux2024}
Black~Forest Labs.
\newblock Flux.
\newblock \url{https://github.com/black-forest-labs/flux}, 2024.

\bibitem[Li et~al.(2022)Li, Li, Xiong, and Hoi]{li2022blip}
Junnan Li, Dongxu Li, Caiming Xiong, and Steven Hoi.
\newblock Blip: Bootstrapping language-image pre-training for unified vision-language understanding and generation.
\newblock In \emph{International conference on machine learning}, pages 12888--12900. PMLR, 2022.

\bibitem[Liu et~al.(2023{\natexlab{a}})Liu, Wu, Van~Hoorick, Tokmakov, Zakharov, and Vondrick]{liu2023zero}
Ruoshi Liu, Rundi Wu, Basile Van~Hoorick, Pavel Tokmakov, Sergey Zakharov, and Carl Vondrick.
\newblock Zero-1-to-3: Zero-shot one image to 3d object.
\newblock In \emph{Proceedings of the IEEE/CVF international conference on computer vision}, pages 9298--9309, 2023{\natexlab{a}}.

\bibitem[Liu et~al.(2023{\natexlab{b}})Liu, Lin, Zeng, Long, Liu, Komura, and Wang]{liu2023syncdreamer}
Yuan Liu, Cheng Lin, Zijiao Zeng, Xiaoxiao Long, Lingjie Liu, Taku Komura, and Wenping Wang.
\newblock Syncdreamer: Generating multiview-consistent images from a single-view image.
\newblock \emph{arXiv preprint arXiv:2309.03453}, 2023{\natexlab{b}}.

\bibitem[Luo et~al.(2023)Luo, Dunlap, Park, Holynski, and Darrell]{luo2023diffusion}
Grace Luo, Lisa Dunlap, Dong~Huk Park, Aleksander Holynski, and Trevor Darrell.
\newblock Diffusion hyperfeatures: Searching through time and space for semantic correspondence.
\newblock \emph{Advances in Neural Information Processing Systems}, 36:\penalty0 47500--47510, 2023.

\bibitem[Maggio et~al.(2023)Maggio, Abate, Shi, Mario, and Carlone]{maggio2023loc}
Dominic Maggio, Marcus Abate, Jingnan Shi, Courtney Mario, and Luca Carlone.
\newblock Loc-nerf: Monte carlo localization using neural radiance fields.
\newblock In \emph{2023 IEEE International Conference on Robotics and Automation (ICRA)}, pages 4018--4025. IEEE, 2023.

\bibitem[Meng et~al.(2024)Meng, Xu, Wang, Cao, and Huang]{meng2024diffusionmodelactivationsevaluated}
Benyuan Meng, Qianqian Xu, Zitai Wang, Xiaochun Cao, and Qingming Huang.
\newblock Not all diffusion model activations have been evaluated as discriminative features.
\newblock In \emph{Annual Conference on Neural Information Processing Systems}, pages 55141--55177, 2024.

\bibitem[Metzer et~al.(2023)Metzer, Richardson, Patashnik, Giryes, and Cohen-Or]{metzer2023latent}
Gal Metzer, Elad Richardson, Or Patashnik, Raja Giryes, and Daniel Cohen-Or.
\newblock Latent-nerf for shape-guided generation of 3d shapes and textures.
\newblock In \emph{Proceedings of the IEEE/CVF conference on computer vision and pattern recognition}, pages 12663--12673, 2023.

\bibitem[Mildenhall et~al.(2021)Mildenhall, Srinivasan, Tancik, Barron, Ramamoorthi, and Ng]{mildenhall2021nerf}
Ben Mildenhall, Pratul~P Srinivasan, Matthew Tancik, Jonathan~T Barron, Ravi Ramamoorthi, and Ren Ng.
\newblock Nerf: Representing scenes as neural radiance fields for view synthesis.
\newblock \emph{Communications of the ACM}, 65\penalty0 (1):\penalty0 99--106, 2021.

\bibitem[Oquab et~al.(2023)Oquab, Darcet, Moutakanni, Vo, Szafraniec, Khalidov, Fernandez, Haziza, Massa, El-Nouby, et~al.]{oquab2023dinov2}
Maxime Oquab, Timoth{\'e}e Darcet, Th{\'e}o Moutakanni, Huy Vo, Marc Szafraniec, Vasil Khalidov, Pierre Fernandez, Daniel Haziza, Francisco Massa, Alaaeldin El-Nouby, et~al.
\newblock Dinov2: Learning robust visual features without supervision.
\newblock \emph{arXiv preprint arXiv:2304.07193}, 2023.

\bibitem[Poole et~al.(2022)Poole, Jain, Barron, and Mildenhall]{poole2022dreamfusion}
Ben Poole, Ajay Jain, Jonathan~T Barron, and Ben Mildenhall.
\newblock Dreamfusion: Text-to-3d using 2d diffusion.
\newblock \emph{arXiv preprint arXiv:2209.14988}, 2022.

\bibitem[Radford et~al.(2021)Radford, Kim, Hallacy, Ramesh, Goh, Agarwal, Sastry, Askell, Mishkin, Clark, et~al.]{radford2021learning}
Alec Radford, Jong~Wook Kim, Chris Hallacy, Aditya Ramesh, Gabriel Goh, Sandhini Agarwal, Girish Sastry, Amanda Askell, Pamela Mishkin, Jack Clark, et~al.
\newblock Learning transferable visual models from natural language supervision.
\newblock In \emph{International conference on machine learning}, pages 8748--8763. PMLR, 2021.

\bibitem[Rombach et~al.(2022)Rombach, Blattmann, Lorenz, Esser, and Ommer]{rombach2022high}
Robin Rombach, Andreas Blattmann, Dominik Lorenz, Patrick Esser, and Bj{\"o}rn Ommer.
\newblock High-resolution image synthesis with latent diffusion models.
\newblock In \emph{Proceedings of the IEEE/CVF conference on computer vision and pattern recognition}, pages 10684--10695, 2022.

\bibitem[Salimans et~al.(2016)Salimans, Goodfellow, Zaremba, Cheung, Radford, and Chen]{salimans2016improved}
Tim Salimans, Ian Goodfellow, Wojciech Zaremba, Vicki Cheung, Alec Radford, and Xi Chen.
\newblock Improved techniques for training gans.
\newblock \emph{Advances in neural information processing systems}, 29, 2016.

\bibitem[Sobol et~al.(2024)Sobol, Xu, and Litany]{sobol2024zero}
Ido Sobol, Chenfeng Xu, and Or Litany.
\newblock Zero-to-hero: Enhancing zero-shot novel view synthesis via attention map filtering.
\newblock \emph{arXiv preprint arXiv:2405.18677}, 2024.

\bibitem[Song et~al.(2020)Song, Meng, and Ermon]{song2020denoising}
Jiaming Song, Chenlin Meng, and Stefano Ermon.
\newblock Denoising diffusion implicit models.
\newblock \emph{arXiv preprint arXiv:2010.02502}, 2020.

\bibitem[Stojanov et~al.(2021)Stojanov, Thai, and Rehg]{Stojanov_2021_CVPR}
Stefan Stojanov, Anh Thai, and James~M. Rehg.
\newblock Using shape to categorize: Low-shot learning with an explicit shape bias.
\newblock In \emph{Proceedings of the IEEE/CVF Conference on Computer Vision and Pattern Recognition (CVPR)}, pages 1798--1808, 2021.

\bibitem[Szegedy et~al.(2016)Szegedy, Vanhoucke, Ioffe, Shlens, and Wojna]{Szegedy_2016_CVPR}
Christian Szegedy, Vincent Vanhoucke, Sergey Ioffe, Jon Shlens, and Zbigniew Wojna.
\newblock Rethinking the inception architecture for computer vision.
\newblock In \emph{Proceedings of the IEEE Conference on Computer Vision and Pattern Recognition (CVPR)}, 2016.

\bibitem[Tang et~al.(2023)Tang, Jia, Wang, Phoo, and Hariharan]{tang2023emergent}
Luming Tang, Menglin Jia, Qianqian Wang, Cheng~Perng Phoo, and Bharath Hariharan.
\newblock Emergent correspondence from image diffusion.
\newblock \emph{Advances in Neural Information Processing Systems}, 36:\penalty0 1363--1389, 2023.

\bibitem[Wang et~al.(2004)Wang, Bovik, Sheikh, and Simoncelli]{wang2004image}
Zhou Wang, Alan~C Bovik, Hamid~R Sheikh, and Eero~P Simoncelli.
\newblock Image quality assessment: from error visibility to structural similarity.
\newblock \emph{IEEE transactions on image processing}, 13\penalty0 (4):\penalty0 600--612, 2004.

\bibitem[Watson et~al.(2022)Watson, Chan, Martin-Brualla, Ho, Tagliasacchi, and Norouzi]{watson2022novel}
Daniel Watson, William Chan, Ricardo Martin-Brualla, Jonathan Ho, Andrea Tagliasacchi, and Mohammad Norouzi.
\newblock Novel view synthesis with diffusion models.
\newblock \emph{arXiv preprint arXiv:2210.04628}, 2022.

\bibitem[Woo et~al.(2024)Woo, Park, Go, Kim, and Kim]{woo2024harmonyview}
Sangmin Woo, Byeongjun Park, Hyojun Go, Jin-Young Kim, and Changick Kim.
\newblock Harmonyview: Harmonizing consistency and diversity in one-image-to-3d.
\newblock In \emph{Proceedings of the IEEE/CVF Conference on Computer Vision and Pattern Recognition}, pages 10574--10584, 2024.

\bibitem[Wu et~al.(2023)Wu, Zhang, Fu, Wang, Ren, Pan, Wu, Yang, Wang, Qian, et~al.]{wu2023omniobject3d}
Tong Wu, Jiarui Zhang, Xiao Fu, Yuxin Wang, Jiawei Ren, Liang Pan, Wayne Wu, Lei Yang, Jiaqi Wang, Chen Qian, et~al.
\newblock Omniobject3d: Large-vocabulary 3d object dataset for realistic perception, reconstruction and generation.
\newblock In \emph{Proceedings of the IEEE/CVF Conference on Computer Vision and Pattern Recognition}, pages 803--814, 2023.

\bibitem[Xiang et~al.(2024)Xiang, Lv, Xu, Deng, Wang, Zhang, Chen, Tong, and Yang]{xiang2024structured}
Jianfeng Xiang, Zelong Lv, Sicheng Xu, Yu Deng, Ruicheng Wang, Bowen Zhang, Dong Chen, Xin Tong, and Jiaolong Yang.
\newblock Structured 3d latents for scalable and versatile 3d generation.
\newblock \emph{arXiv preprint arXiv:2412.01506}, 2024.

\bibitem[Xu et~al.(2024)Xu, Ling, Fidler, and Litany]{xu20243difftection}
Chenfeng Xu, Huan Ling, Sanja Fidler, and Or Litany.
\newblock 3difftection: 3d object detection with geometry-aware diffusion features.
\newblock In \emph{Proceedings of the IEEE/CVF Conference on Computer Vision and Pattern Recognition}, pages 10617--10627, 2024.

\bibitem[Yu et~al.(2023)Yu, Forghani, Derpanis, and Brubaker]{yu2023long}
Jason~J Yu, Fereshteh Forghani, Konstantinos~G Derpanis, and Marcus~A Brubaker.
\newblock Long-term photometric consistent novel view synthesis with diffusion models.
\newblock In \emph{Proceedings of the IEEE/CVF International Conference on Computer Vision}, pages 7094--7104, 2023.

\bibitem[Zhan et~al.(2025)Zhan, Zheng, Xie, and Zisserman]{zhan2025general}
Guanqi Zhan, Chuanxia Zheng, Weidi Xie, and Andrew Zisserman.
\newblock A general protocol to probe large vision models for 3d physical understanding.
\newblock \emph{Advances in Neural Information Processing Systems}, 37:\penalty0 43468--43498, 2025.

\bibitem[Zhang et~al.(2024)Zhang, Wang, Zhang, Qiu, Pang, Jiang, Yang, Xu, and Yu]{zhang2024clay}
Longwen Zhang, Ziyu Wang, Qixuan Zhang, Qiwei Qiu, Anqi Pang, Haoran Jiang, Wei Yang, Lan Xu, and Jingyi Yu.
\newblock Clay: A controllable large-scale generative model for creating high-quality 3d assets.
\newblock \emph{ACM Transactions on Graphics (TOG)}, 43\penalty0 (4):\penalty0 1--20, 2024.

\bibitem[Zhang et~al.(2018)Zhang, Isola, Efros, Shechtman, and Wang]{zhang2018unreasonable}
Richard Zhang, Phillip Isola, Alexei~A Efros, Eli Shechtman, and Oliver Wang.
\newblock The unreasonable effectiveness of deep features as a perceptual metric.
\newblock In \emph{Proceedings of the IEEE conference on computer vision and pattern recognition}, pages 586--595, 2018.

\bibitem[Zhou et~al.(2025)Zhou, Gao, Voleti, Vasishta, Yao, Boss, Torr, Rupprecht, and Jampani]{zhou2025stable}
Jensen~Jinghao Zhou, Hang Gao, Vikram Voleti, Aaryaman Vasishta, Chun-Han Yao, Mark Boss, Philip Torr, Christian Rupprecht, and Varun Jampani.
\newblock Stable virtual camera: Generative view synthesis with diffusion models.
\newblock \emph{arXiv preprint arXiv:2503.14489}, 2025.

\end{thebibliography}
}
\clearpage
\setcounter{page}{1}
\maketitlesupplementary

\section{Training Hyper-Parameter Summary}
\label{app:hyperparams}

\begin{table}[h]
\centering
\small
\caption{Summary of training hyper-parameters.}
\label{tab:hyperparams}
\resizebox{\linewidth}{!}{
\begin{tabular}{ll}
\toprule
\textbf{Projection Head} & 2-layer MLP, ReLU, output dim = 2048, $\ell_2$-norm \\
\textbf{Optimizer} & AdamW, lr=$1\!\times\!10^{-4}$, batch size=32, epochs=100 \\
\textbf{Loss} & Triplet loss, margin $m=1.0$, negatives weighted by mask size \\
\bottomrule
\end{tabular}}
\end{table}

\section{Dataset and Benchmarks}
\label{app:dataset}

\subsection{Mask Generation Procedures}
\label{app:mask_generation}

\paragraph{View grid.}  
We render a discrete grid of 16 azimuths (0°–337.5° in 22.5° steps), elevation 30°, and camera distance 2.7. Rasterization in PyTorch3D uses $512\times512$ resolution, faces-per-pixel $=1$, blur radius $=0$, and a \texttt{SoftPhongShader} with ambient lights. Background masks are derived from the RGBA alpha channel. All RGB object renderings used in the inpainting tasks were produced in \textbf{Blender~4.4} with the same camera parameters, using the EEVEE renderer.

\paragraph{Visibility and invisibility masks.}  
Source-visible triangles are identified with a z-buffer depth test in the source view and reprojected to the target to form a binary visibility mask; the complement within the object silhouette yields the invisibility mask. Both masks are regularized with OpenCV morphological operations: closing (elliptical kernel radius 4), removal of overlaps, then opening (radius 10). This produces clean boundaries and disjoint support.

\paragraph{Epipolar masks and neighborhood constraint.}  
For each source pixel, we trace a line from the source camera center through its corresponding visible surface point, reprojected into the target view. Segments that remain unoccluded are projected to the target image, yielding a sparse epipolar mask. We regularize this mask with closing (elliptical kernel radius 3) and further constrain it by intersecting with a dilated object silhouette (kernel radius 20). The resulting regions extend along realistic object contours, ensuring that edits occur in occluded volumes adjacent to the object rather than drifting into background areas.

\begin{figure}[h]
    \centering
    \includegraphics[width=\linewidth]{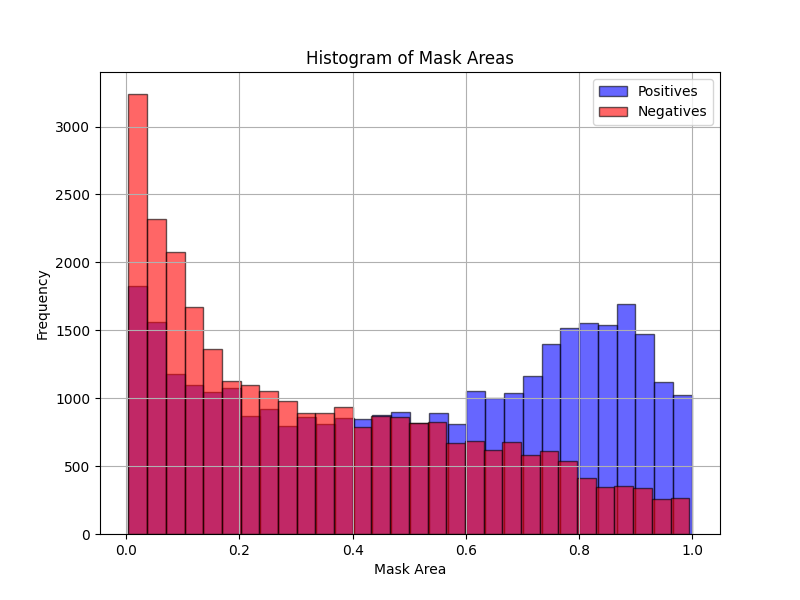}
    \caption{Histogram of raw mask areas before adding epipolar regions. 
    Invisibility masks (blue) are distributed toward larger areas, while visibility masks (red) are concentrated at smaller values.}
    \label{fig:mask_histogram}
\end{figure}

\paragraph{Positive and negative masks.}  
Final training masks are constructed as:
\begin{itemize}[leftmargin=*]
    \item Positives (plausible): invisibility $\cup$ epipolar,
    \item Negatives (implausible): visibility $\cup$ epipolar.
\end{itemize}
Before adding epipolar regions, invisibility masks cover larger areas than visibility masks (see \cref{fig:mask_histogram}). 
This imbalance is a technical artifact and is corrected during training by weighting samples according to mask area.

\subsection{Inpainting Settings}  
\begin{figure*}[h!]
    \centering
    \includegraphics[width=\linewidth]{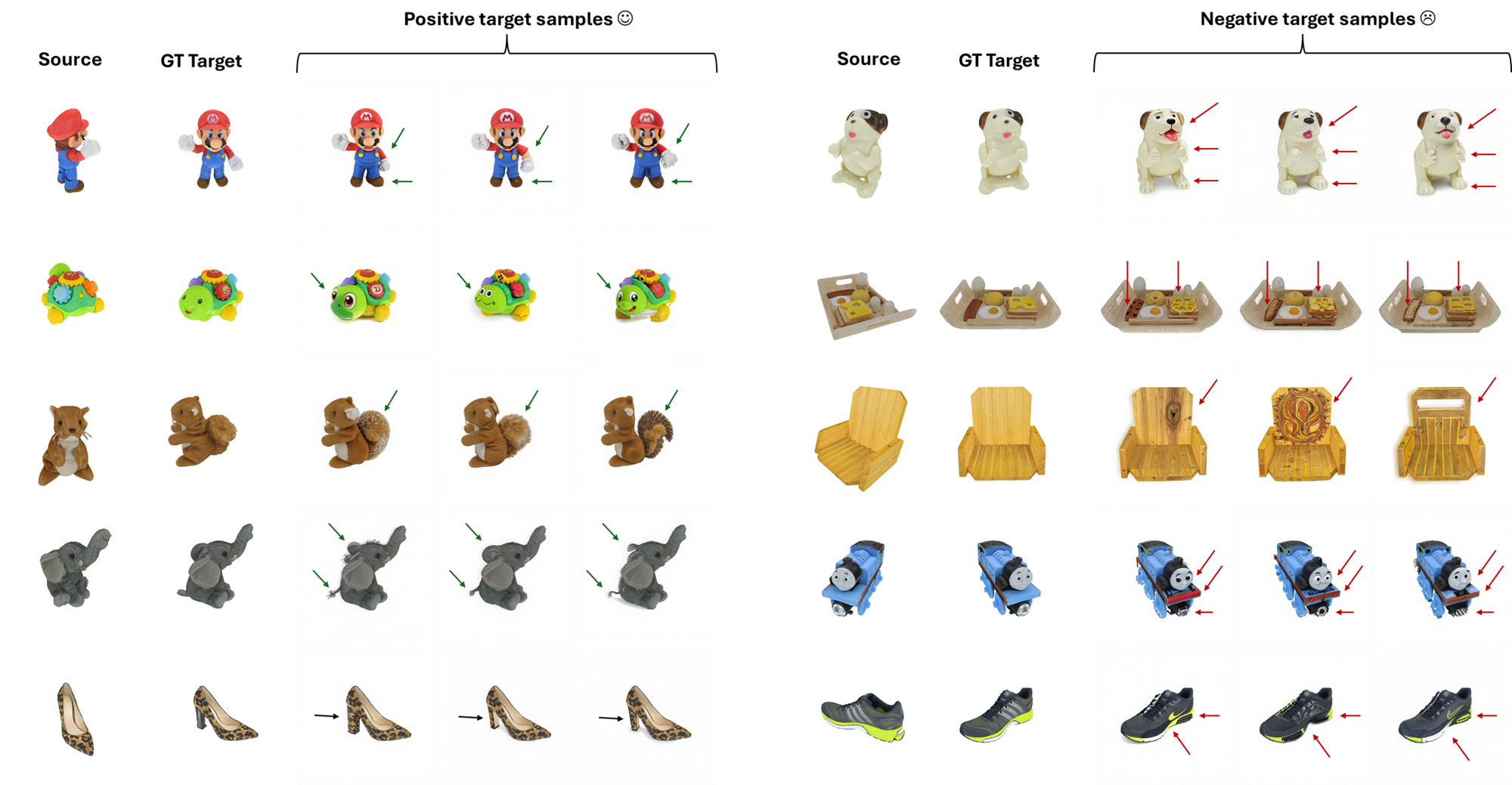}
    \caption{Additional examples from our dataset, showing source views, target views, and corresponding positive and negative inpainted samples.}
    \label{fig:dataset_full}
\end{figure*}

We use the FLUX ControlNet inpainting backbone\footnote{\texttt{alimama-creative/FLUX.1-dev-Controlnet-Inpainting-Beta}} with fused LoRA\footnote{\texttt{alimama-creative/FLUX.1-Turbo-Alpha}}, captions from BLIP, and a fixed seed (45). Control images are target renders over white, resized with LANCZOS to $768\times768$; control masks are resized with NEAREST. Parameters: inpainting strength 0.9, 8 steps, guidance scale 3.5, true guidance 1.0. Prompts enforce semantic plausibility, with the following negative prompt: \emph{``Blurred details, low res, unrealistic textures, dark or cluttered backgrounds, dull colors, distorted shapes''}. Small-mask cases are automatically filtered during training.  
This protocol yields exactly $40 \times 16 \times 15 \times 4$ positive examples, with an equal number of negatives. The final \datasetname benchmark is split into 32 training objects and 8 test objects, disjoint from the 40 objects used in the user study. Additional positive/negative examples from \datasetname are provided in \cref{fig:dataset_full}.

\subsection{Misaligned Views Protocol}
\label{app:misaligned}

The \emph{Misaligned-GSO} benchmark tests whether a metric is sensitive to the \emph{intended} target view rather than to any other plausible ground-truth view of the same object. 
For each source–target pair, we replace the target with an alternative rendering sampled from our 16 uniformly spaced azimuths (22.5° steps, elevation 30°, distance 2.7), yielding 15 misaligned alternatives per source. 
The most severe misalignment corresponds to the opposite $180^\circ$ view. 

All misaligned views are true Blender~4.4/EEVEE renderings under identical camera settings (not inpainted), ensuring high visual quality. 
They depict the correct object at the wrong target angle, isolating viewpoint inconsistency. 
When used in contrastive training (\cref{app:appendix-contrastive}), these misaligned views serve as additional negatives, complementing the implausible inpainted views of \datasetname.

\section{Backbones and Feature Extractors}
\label{app:models}

\subsection{U-Net Architecture in Stable Diffusion and Its Connection to NVS Models}
\label{app:appendix-unet}

\paragraph{U-Net backbone in Stable Diffusion.}
The Zero123 and Zero123-XL models we employ for feature extraction are fine-tuned variants of Stable Diffusion (SD)~\cite{rombach2022high}, a latent diffusion framework originally developed for text-to-image generation. The SD denoiser is a U-Net with a hierarchical encoder–decoder topology comprising nine stages: four downsampling encoder blocks, a central bottleneck, and four upsampling decoder blocks. Each block integrates ResNet layers, self-attention, and (for conditional tasks) cross-attention. Spatial resolution is reduced by a factor of two at each encoder stage and symmetrically recovered in the decoder.

\paragraph{Adaptation to NVS models.}
Zero123-XL inherits this structure but replaces text conditioning with conditioning on a source image and a relative viewpoint specification, enabling novel view synthesis. This preserves the same multi-scale hierarchy:
\begin{itemize}
    \item \emph{Early encoder layers} emphasize fine appearance cues.
    \item \emph{Bottleneck features} capture coarse global structure.
    \item \emph{Decoder layers} upsample features and output the noise prediction used by the diffusion process.
\end{itemize}

\paragraph{Feature extraction protocol.}
Features can be extracted at different abstraction levels within this U-Net. Following prior work~\cite{meng2024diffusionmodelactivationsevaluated}, we run the denoiser for a single step at a specific noise level and record activations. While higher timesteps emphasize coarse geometry and lower ones preserve finer details, we found that $t=0$ provides stable and informative features (see~\cref{tab:combined_metrics}). Inspired by HyperFeatures~\cite{luo2023diffusion}, which emphasize the benefit of multi-scale representations, we extract features from the outputs of all nine U-Net blocks, spanning low-level appearance to high-level structure.

\subsection{Baseline Feature Extractors}
\label{app:appendix-baselines}

We evaluate standard vision backbones as baselines: InceptionV3~\cite{Szegedy_2016_CVPR}, CLIP (\texttt{ViT-L/14@336px})~\cite{radford2021learning}, and DINOv2 (\texttt{ViT-L/14})~\cite{oquab2023dinov2}.  
In addition to plain features, we form simple concatenation baselines by stacking source and target embeddings, with or without the relative viewpoint angle. For example, CLIP-CAT is just the concatenation of source and target CLIP features, while CLIP-CAT-Angle appends the angle as an additional input; the same variants are defined for Inception and DINOv2.

For distribution-level metrics, we adopt the standard pairings from prior work: Fréchet distance on Inception (FID) and its joint variant JFID~\cite{elata2024novel}, Fréchet distance on DINOv2 (FDD) and JFDD~\cite{elata2024novel}, and maximum mean discrepancy on CLIP (CMMD)~\cite{jayasumana2024rethinking}. We also introduce a new joint variant, JCMMD, which applies MMD to CLIP-CAT features. This setup lets us test whether generic vision embeddings, with or without trivial concatenation, can meaningfully serve as NVS evaluation metrics.

\section{Feature Adaptation by Contrastive Fine-Tuning}
\label{app:appendix-contrastive}

The raw diffusion features $v$ encode multi-scale cues from a source--pose--target triplet, but they are not explicitly optimized to distinguish plausible from implausible generations. To specialize them for evaluation, we train a lightweight projection network with a triplet loss on \datasetname.

\paragraph{Training protocol.}
For each ground-truth triplet $(I_{\text{src}}, I_{\text{tgt}}, \pi)$ we form contrastive examples:
\begin{itemize}
    \item \textbf{Positive:} one of four plausible inpainted completions consistent with the source and pose.
    \item \textbf{Negative:} either (i) an implausible inpainted sample from \datasetname, or (ii) a misaligned-angle target rendering.
\end{itemize}
This yields multiple triplets per anchor. The embedding is optimized with a margin-based triplet loss:
\begin{equation}
    \mathcal{L}(a,p,n) = \max \big( \|a-p\|_2 - \|a-n\|_2 + m, \, 0 \big),
\end{equation}
with margin $m=1.0$. During training, negatives are downweighted according to their mask size relative to the object silhouette: very small masks contribute negligible loss and are effectively skipped. Positives are always retained.

\paragraph{Lightweight projection.}
The network is a two-layer MLP with ReLU activation that reduces feature dimensionality approximately by a factor of four and applies $\ell_2$ normalization:
\begin{equation}
    f_{\text{PRISM}} = \frac{h(v)}{\|h(v)\|}.
\end{equation}
This compact embedding (2048-dim) maintains discriminative power while being efficient for distributional metrics such as MMD.

We train for 100 epochs with AdamW (learning rate $10^{-4}$), batch size 32, and early stopping based on validation loss.

\section{Extended Experiments}
\label{app:exp}

\subsection{Robustness to Image Degradations}
\label{app:image_degradations}
\begin{figure}[t]
  \centering
  \includegraphics[width=\linewidth]{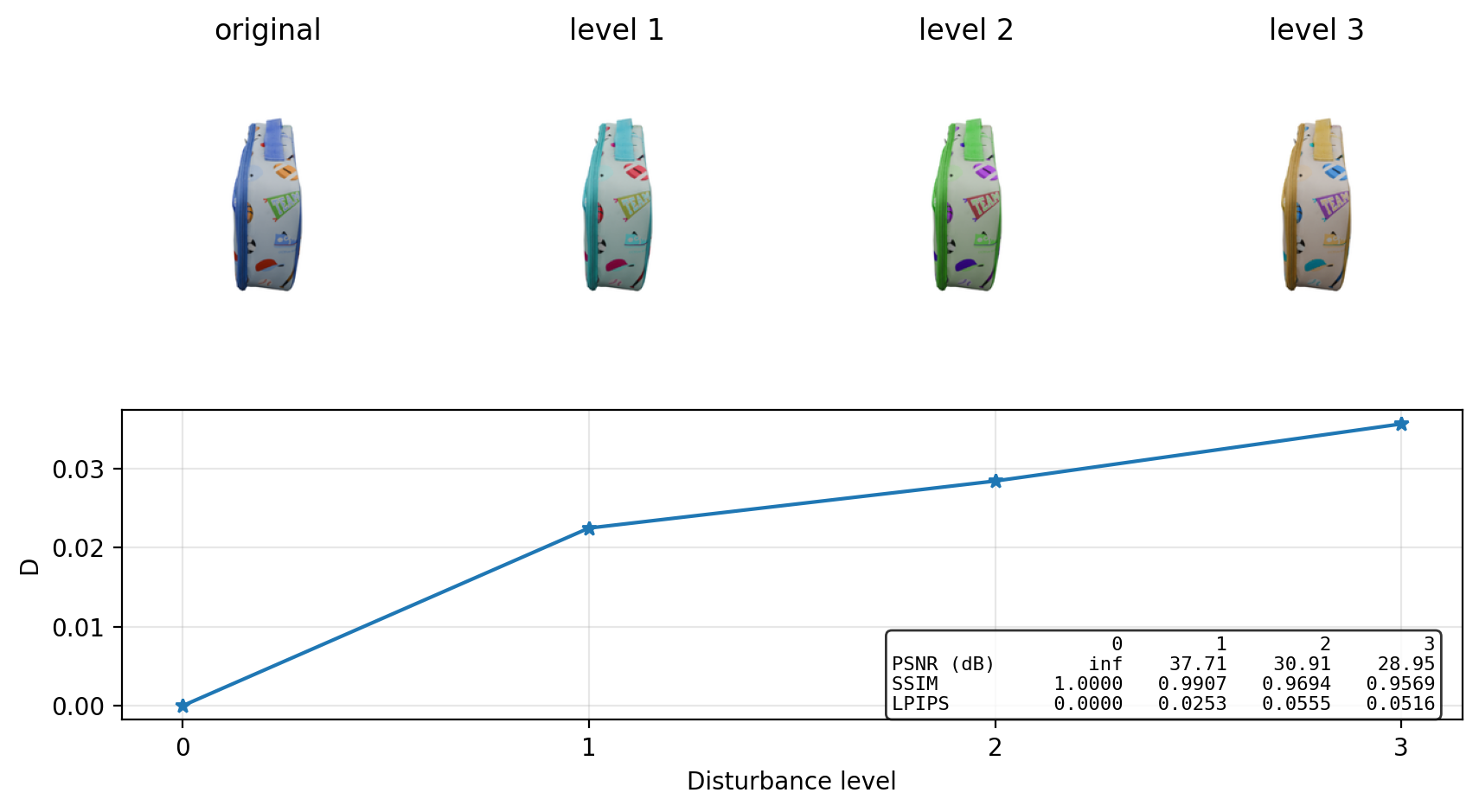}
  \caption{Degradation under color shifts at increasing intensity levels. The $y$-axis reports $D_{\text{PRISM}}$}
  \label{fig:color}
\end{figure}

\begin{figure}[t]
  \centering
  \includegraphics[width=\linewidth]{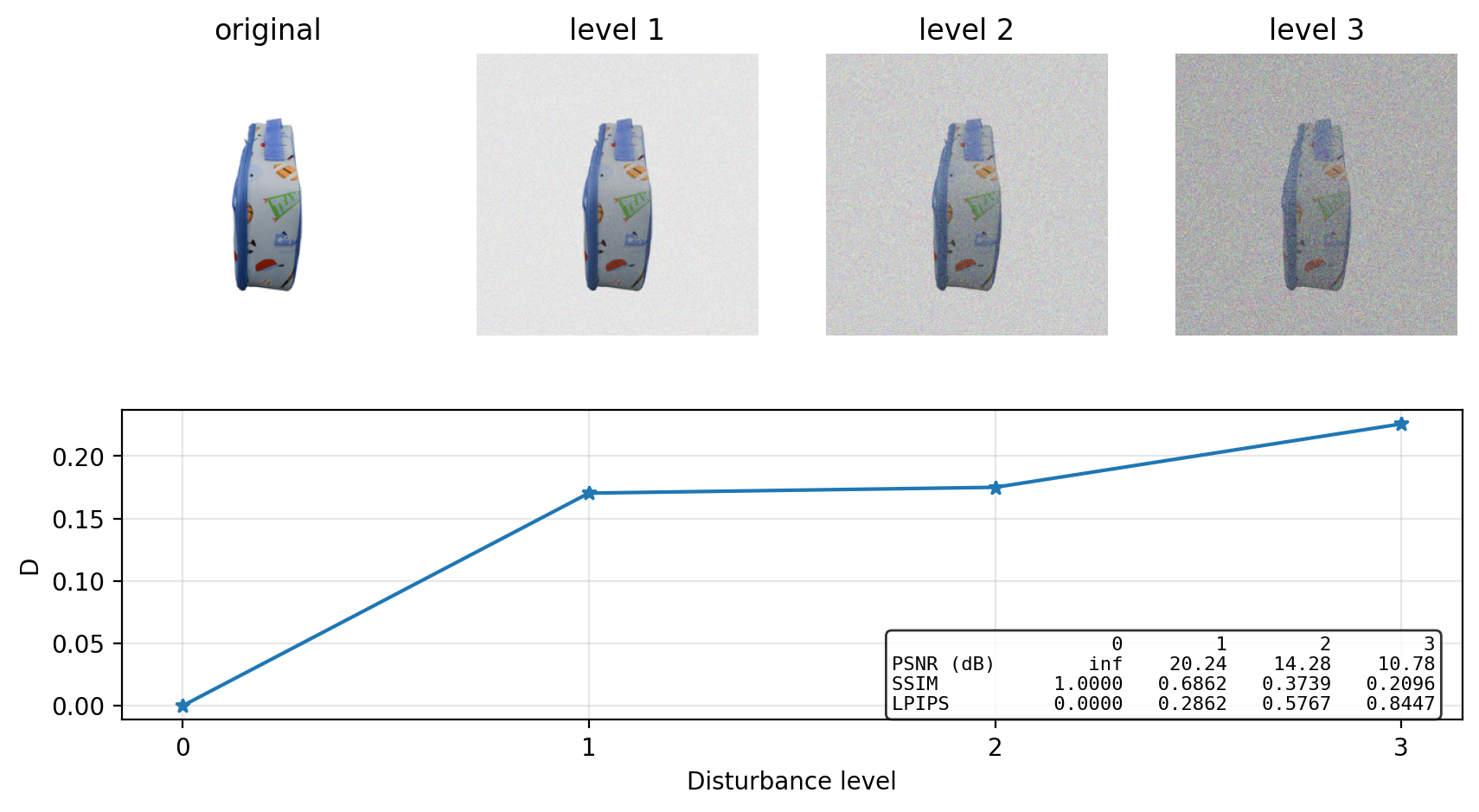}
  \caption{Degradation under Gaussian noise at increasing intensity levels. The $y$-axis reports $D_{\text{PRISM}}$}
  \label{fig:noise}
\end{figure}

\begin{figure}[t]
  \centering
  \includegraphics[width=\linewidth]{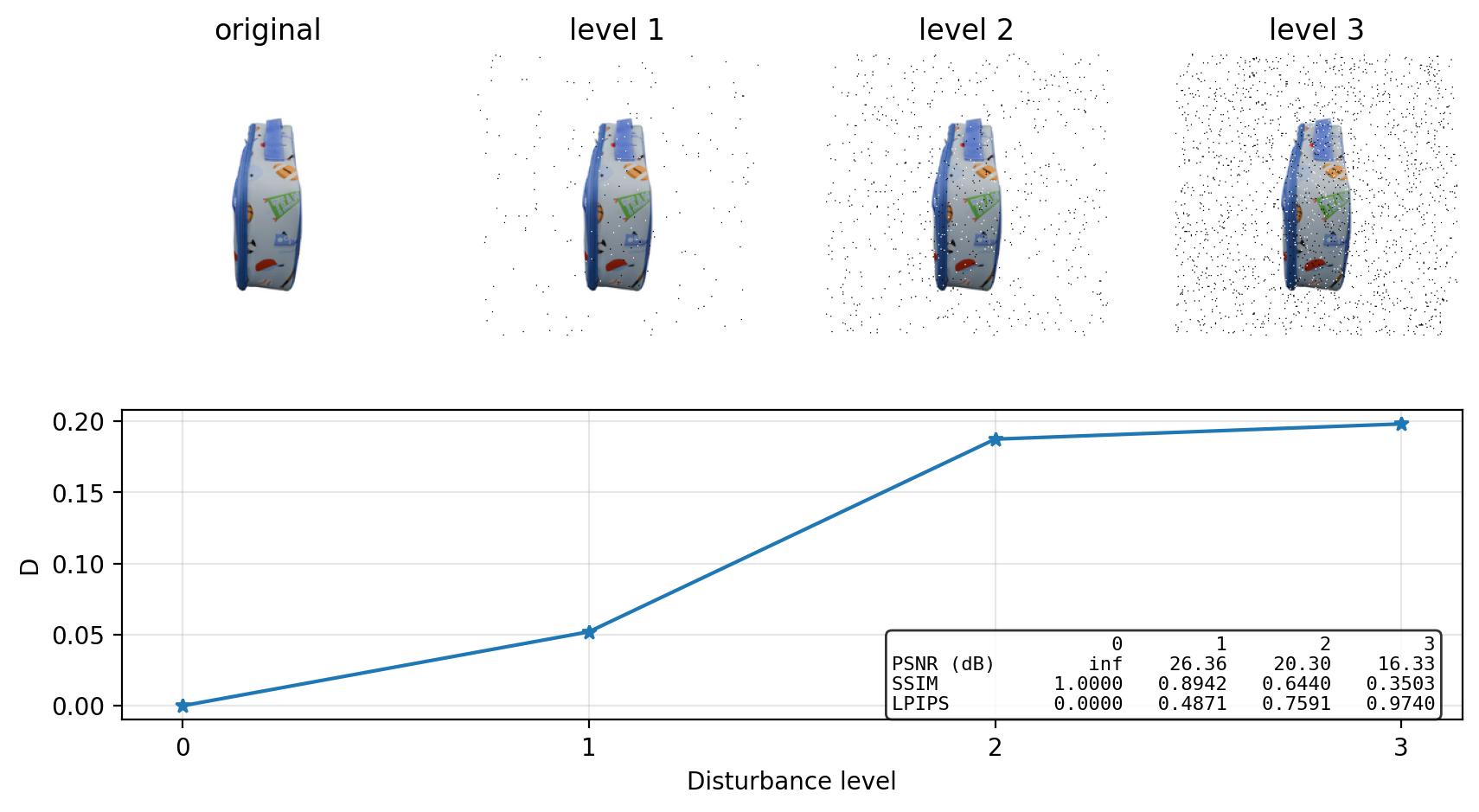}
  \caption{Degradation under salt-and-pepper noise at increasing intensity levels. The $y$-axis reports $D_{\text{PRISM}}$}
  \label{fig:saltpepper}
\end{figure}

We evaluate sensitivity to four corruptions applied only to the target: Gaussian blur, hue shifts, additive Gaussian noise, and salt--pepper noise. 
All images are composited to white and resized to $256{\times}256$, and metrics (PSNR, SSIM, LPIPS, $D_{\text{PRISM}}$) are computed on $[0,1]$ tensors. 
Each corruption is tested at three severity levels on 200 source--target pairs. 
Blur radii are $\{1.0, 1.5, 5.0\}$, hue shifts $\{-0.1, -0.3, -0.5\}$, Gaussian noise is blended with weights $t=\{0.8, 0.6, 0.4\}$, and salt--pepper flip rates are $\{0.005, 0.02, 0.05\}$. 
As shown in \cref{fig:blur_main,fig:color,fig:noise,fig:saltpepper}, standard metrics degrade monotonically with severity, and $D_{\text{PRISM}}$ follows the same trend, confirming robustness to corruption.

\subsection{Sensitivity to Pose Misalignment}
\begin{figure}[h!]
    \centering
    \includegraphics[width=1\linewidth]{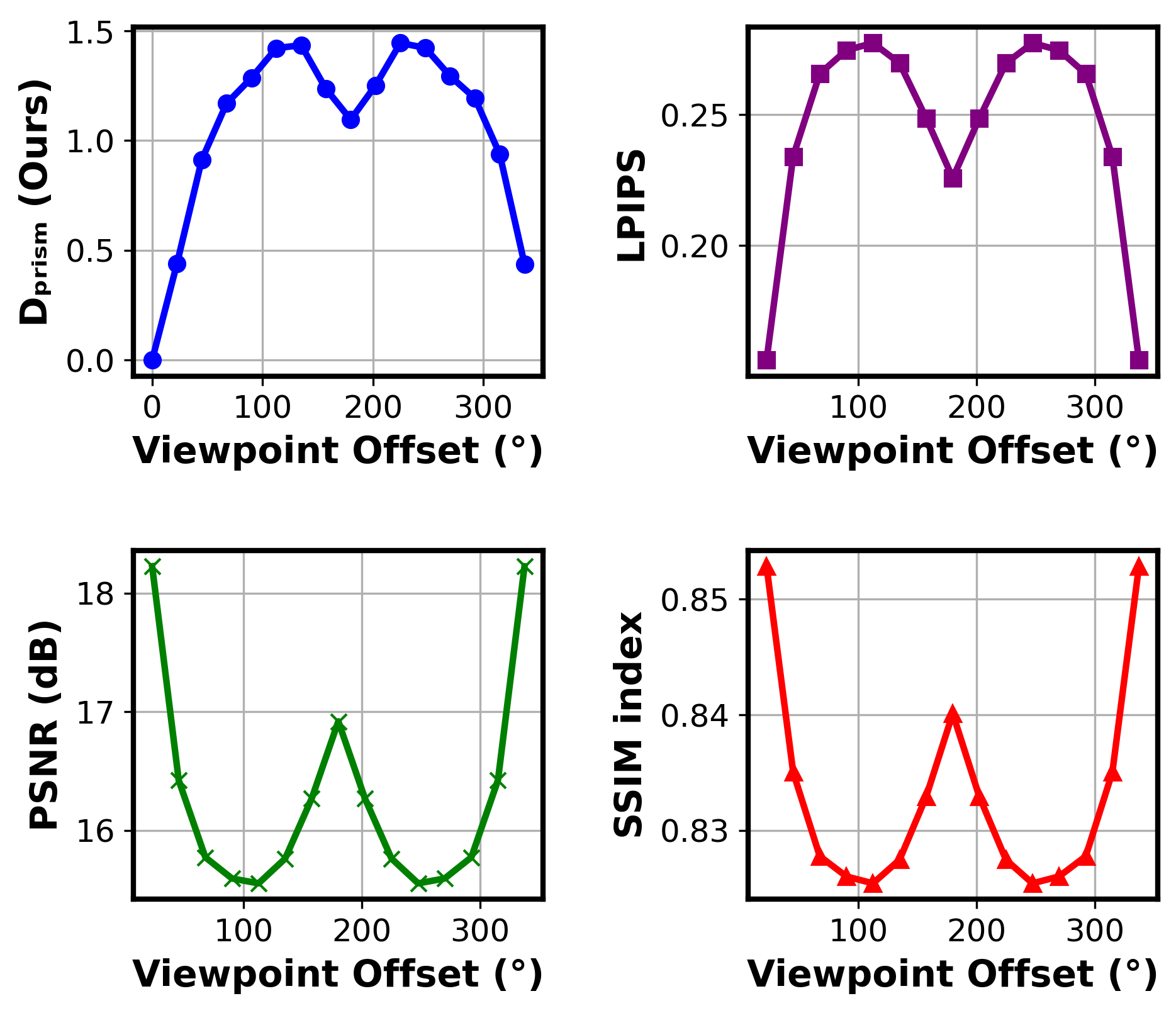}
    \caption{Full-reference evaluation on  misaligned-views.}
    \label{fig:pose_full_ref}
\end{figure}

We also evaluate on the \emph{misaligned-views} benchmark (\cref{app:misaligned}), where targets are replaced by ground-truth renderings at azimuth offsets up to $337.5^\circ$ in $22.5^\circ$ increments. 
This is the full-reference counterpart to the reference-free experiment in the main paper (\cref{fig:pose_no_ref}). 
\Cref{fig:pose_full_ref} shows an M-shaped error curve: errors grow with increasing offset, ease near $180^\circ$, rise again, and finally return to zero at $360^\circ$. 
This confirms that our metrics are sensitive to geometric alignment rather than appearance alone.

\subsection{Linear Classifier Performance Across Timesteps}

This experiment is the extended version of the analysis in the main
paper, reported here with the full set of timesteps. We train linear
classifiers on diffusion features $v$ extracted at timesteps
$t \in [0,999]$ and measure AUC for distinguishing plausible from
implausible views. As shown in \cref{tab:linear-classifier-full},
separability is highest at $t=0$ (AUC 0.90) and gradually decreases
as $t$ increases, confirming that early denoising steps preserve
discriminative structural information, whereas later steps emphasize
generative refinement.

\begin{table}[h!]
    \centering
    \small
    \setlength{\tabcolsep}{8pt} 
    \caption{AUC of linear classifiers trained on features from different timesteps $v$. Higher is better.}
    \label{tab:linear-classifier-full}
    \begin{tabular}{lc}
        \toprule
        \textbf{Timestep} & \textbf{AUC $\uparrow$} \\ 
        \midrule
        $t=0$   & 0.90 \\ 
        $t=100$ & 0.87 \\ 
        $t=200$ & 0.86 \\ 
        $t=300$ & 0.82 \\ 
        $t=400$ & 0.77 \\ 
        $t=500$ & 0.74 \\ 
        $t=600$ & 0.62 \\ 
        $t=700$ & 0.63 \\ 
        $t=800$ & 0.68 \\ 
        $t=900$ & 0.62 \\ 
        $t=999$ & 0.66 \\ 
        \bottomrule
    \end{tabular}
\end{table}

\subsection{Full Ranking Results}
\label{app:full_ranking}

We evaluate six representative NVS models spanning single-image diffusion, multi-view diffusion, and regression with 3D priors (see Table~\ref{tab:eval_models}). These cover the full spectrum of recent approaches and are used consistently across our ranking benchmarks. For the user study, we selected the five models most widely adopted in prior perceptual evaluations (Zero123, Zero123-XL, SEVA, OpenLRM, TRELLIS).

\begin{table}[h]
\centering
\small
\caption{Evaluated models grouped by paradigm.}
\label{tab:eval_models}
\resizebox{\linewidth}{!}{
\begin{tabular}{lll}
\toprule
\textbf{Model} & \textbf{Paradigm} & \textbf{Notes} \\
\midrule
Zero123~\cite{liu2023zero} & Diffusion (single-view) & Zero-shot, camera-conditioned \\
Zero123-XL~\cite{deitke2023objaverse} & Diffusion (single-view) & Larger ObjaverseXL dataset \\
OpenLRM~\cite{hong2023lrm, openlrm} & Regression (NeRF prior) & Predicts NeRFs from one image \\
SEVA~\cite{zhou2025stable} & Diffusion (camera-trajectory) & Multi-view-consistent rendering \\
TRELLIS~\cite{xiang2024structured} & Diffusion (3D latent) & Latent decoded into NeRF/mesh/3DGS \\
SyncDreamer~\cite{liu2023syncdreamer} & Diffusion (multi-view) & Cross-view synchronized denoising \\
\bottomrule
\end{tabular}}
\end{table}

\Cref{tab:gso_ranking,tab:omniobject3d,tab:toys4k,tab:all_datasets} report the full results of our ranking experiments from~\cref{subsec:nvs_ranking}. These tables expand the summary in the main paper, showing complete scores for all baselines (FID, CMMD, FDD, JFID, JCMMD, JFDD) and for our $\text{MMD}_{\text{PRISM}}$ across GSO, Toys4K, and OmniObject3D. The consistent trends confirm the stability of our metric across datasets of varying difficulty.

\begin{table}
    \centering
    \caption{\textbf{Ranking on GSO dataset.} 
    Comparison of reference-free metrics for ranking NVS models. 
    Lower is better.}
    \label{tab:gso_ranking}
    \small
    \setlength{\tabcolsep}{4pt}
    \resizebox{\linewidth}{!}{
        \begin{tabular}{lcccccc}
        \toprule
        \textbf{Metric} & \textbf{OpenLRM} & \textbf{Z123} & \textbf{Z123-XL} & \textbf{TRELLIS} & \textbf{SEVA} & \textbf{SyncDreamer} \\
        \midrule
        FID    & 122.1751 & 111.7643 & 110.8888 & 110.0913 & 109.6447 & 116.3227 \\
        CMMD   & 146.5829 & 78.1093  & 82.0823  & 20.1113  & 29.1913  & 109.9894 \\
        FDD    & 0.4932   & 0.4569   & 0.4567   & 0.3929   & 0.4413   & 0.5258   \\ 
        \midrule
        JFID   & 280.4828 & 267.0230 & 265.5372 & 266.7810 & 264.6224 & 273.9396 \\
        JCMMD  & 1.2579   & 0.7018   & 0.7367   & 0.1907   & 0.2787   & 0.9201   \\
        JFDD   & 1.0533   & 0.9940   & 0.9973   & 0.9247   & 0.9781   & 1.0864   \\
        \midrule
        $\text{MMD}_{\text{PRISM}}$ (Ours) 
               & 0.8415   & 0.3552   & 0.2997   & 0.2858   & \textbf{0.1231} & 0.1392 \\
        \bottomrule
        \end{tabular}
    }
\end{table}

\begin{table}[t]
    \centering
    \caption{\textbf{Ranking on OmniObject3D dataset.} 
    Comparison of 
    erence-free metrics for NVS models. 
    Lower is better.}
    \label{tab:omniobject3d}
    \small
    \setlength{\tabcolsep}{4pt}
    \resizebox{\linewidth}{!}{
        \begin{tabular}{lcccccc}
        \toprule
        \textbf{Metric} & \textbf{OpenLRM} & \textbf{Z123} & \textbf{Z123-XL} & \textbf{TRELLIS} & \textbf{SEVA} & \textbf{SyncDreamer} \\
        \midrule
        FID    & 159.1310 & 146.6099 & 147.9158 & 152.0511 & 153.1845 & 165.1502 \\
        CMMD   & 107.3039 & 91.3282  & 88.7044  & 36.7132  & 41.9074  & 94.5318  \\
        FDD    & 0.6603   & 0.6426   & 0.6433   & 0.6285   & 0.6665   & 0.7576   \\ 
        \midrule
        JFID   & 340.1759 & 327.2526 & 328.8221 & 333.5065 & 336.4982 & 349.2438 \\
        JCMMD  & 0.7935   & 0.7385   & 0.7172   & 0.3085   & 0.3557   & 0.7303   \\
        JFDD   & 1.3583   & 1.3294   & 1.3282   & 1.3162   & 1.3593   & 1.4678   \\
        \midrule
        $\text{MMD}_{\text{PRISM}}$ (Ours) 
               & 1.3752   & 0.6376   & 0.5206   & 0.5696   & 0.2896   & \textbf{0.2030} \\
        \bottomrule
        \end{tabular}
    }
\end{table}

\section{User Study Protocol and Interface}
\label{app:user_study}

We designed a user study to evaluate how well metrics align with human perception of novel view synthesis (NVS). Each task consisted of a source image, a relative camera transformation, and two generated target views. Participants compared the two views along four aspects: \textit{Viewpoint Accuracy}, \textit{Shared Region Consistency}, \textit{Plausibility of New Regions}, and \textit{Image Quality}. 

\begin{figure*}
    \centering
    \subfloat[Example with mostly unseen regions]{%
        \includegraphics[width=\linewidth]{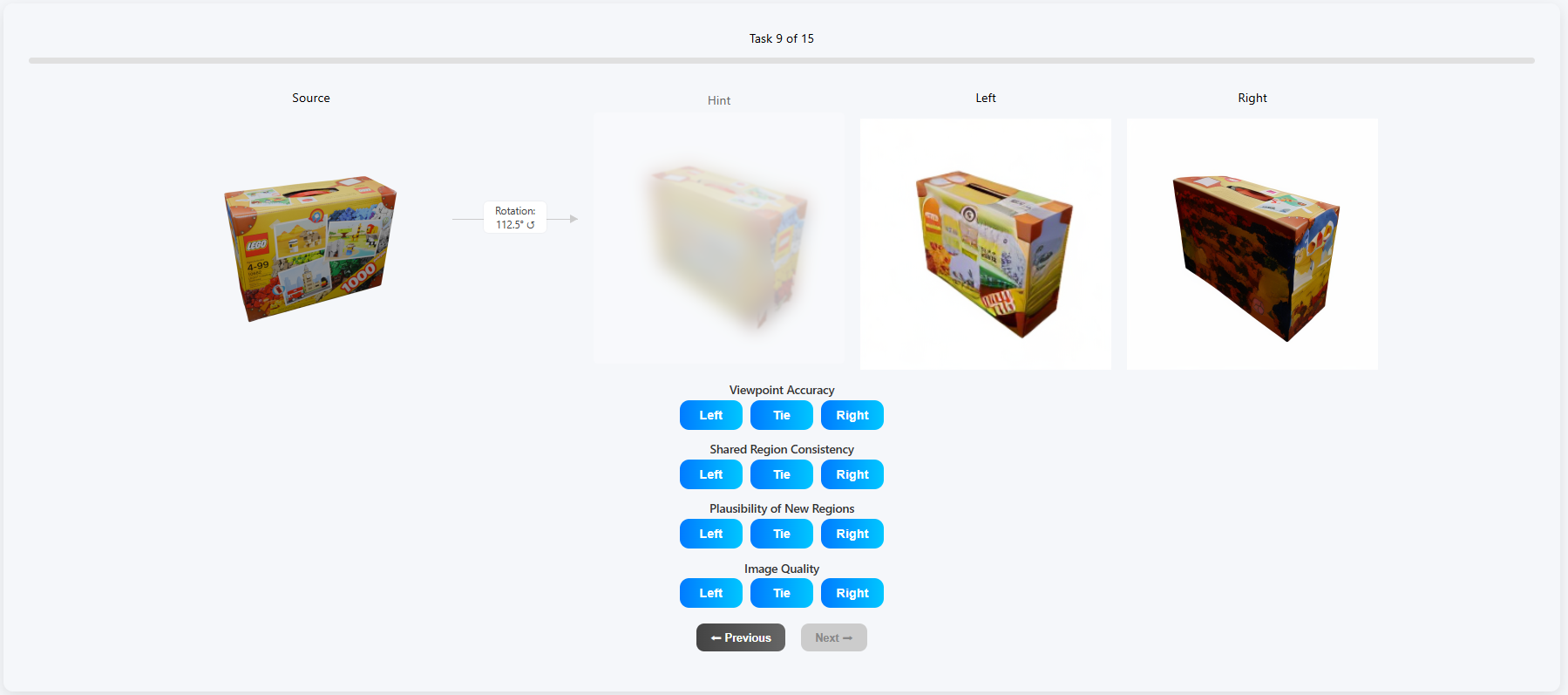}%
        \label{fig:user-study-screenshot-1}
    }
    \hfill
    \subfloat[Example with mostly shared regions]{%
        \includegraphics[width=\linewidth]{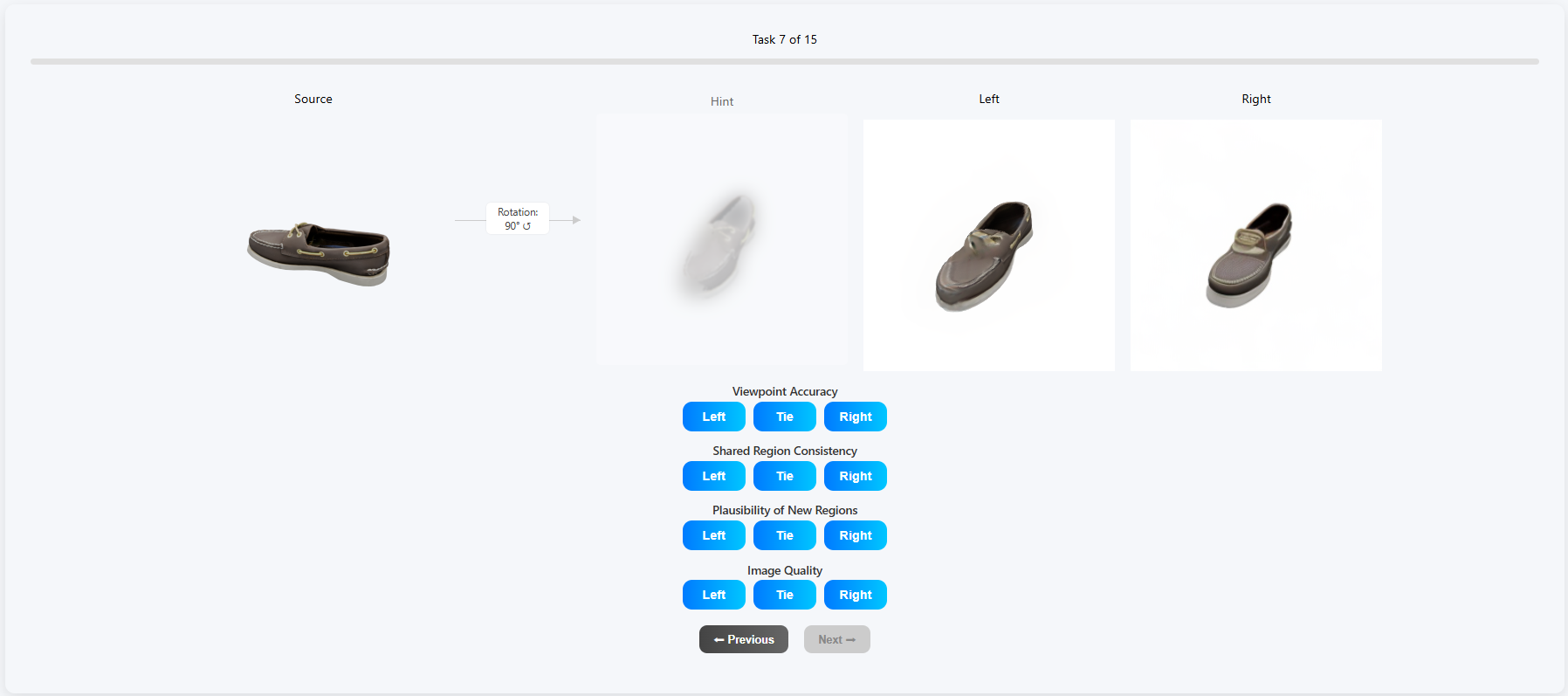}%
        \label{fig:user-study-screenshot-2}
    }
    \caption{Illustrative questions from the user study interface. 
    (a) Source and target views share only a small overlap, so the task emphasizes evaluating \emph{plausibility of new regions}. 
    (b) Source and target views share a large overlap, so the task emphasizes evaluating \emph{shared region consistency}. 
    In both cases, participants rated all four aspects, and the blurred hint provides a uniformly faded reference for geometry without revealing fine detail. Examples were selected with correct viewing angles and good visual quality to represent typical tasks.}
    \label{fig:user-study-interface}
\end{figure*}

To make the evaluation feasible for humans, we provided a blurred hint of the ground-truth target view. Judging large rotations (e.g., $112.5^\circ$) without any reference is not trivial, and the hint served as a coarse anchor while still obscuring fine details. Importantly, regions visible in the source image were preserved to support checking geometric consistency, while unseen regions were heavily blurred. To avoid an unnatural mixture of sharp and blurred patches, we applied a final global blur, producing a uniformly faded image. This way, the hint looks perceptually coherent: it provides structural cues for alignment while withholding pixel-level detail. Examples of the interface are shown in \cref{fig:user-study-screenshot-1,fig:user-study-screenshot-2}.

\begin{figure*}
    \centering
    \includegraphics[width=\linewidth]{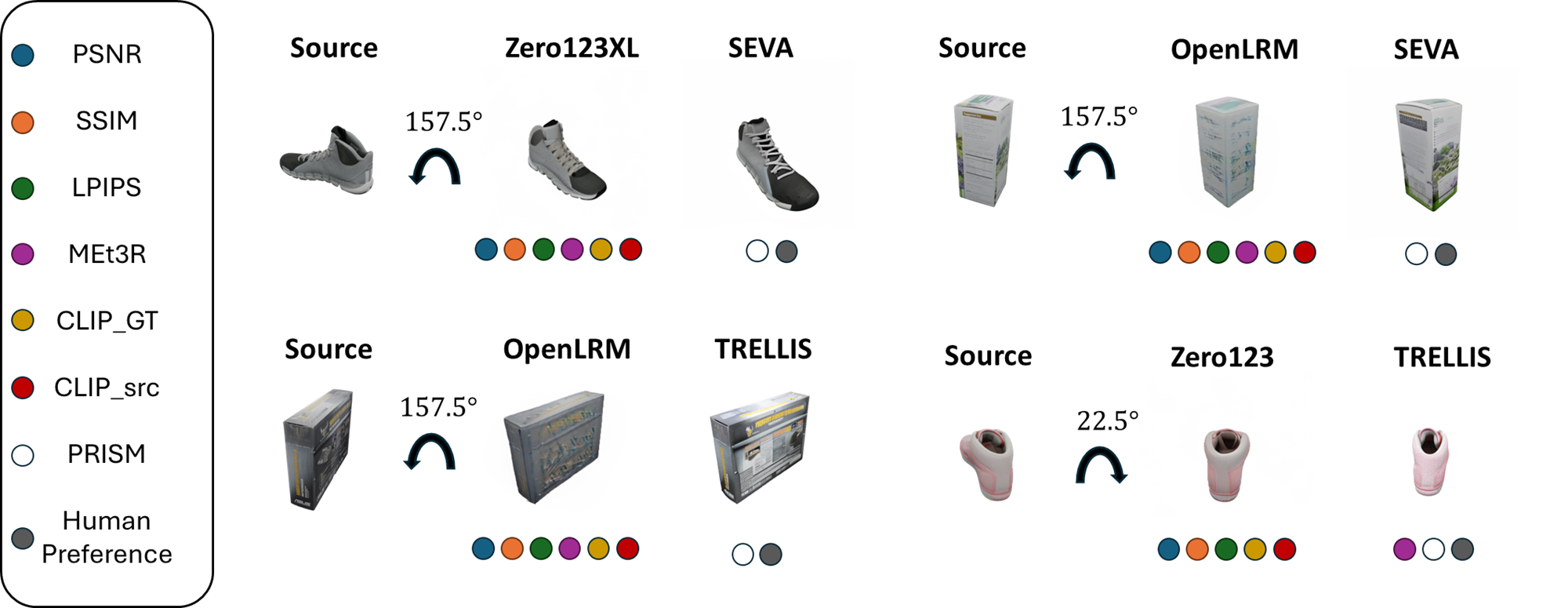}
    \caption{Cases from the user study focused on the aspect of plausibility in newly synthesized regions, where human participants consistently preferred one generated view over another. Standard metrics—including PSNR, SSIM, LPIPS, MEt3R, and CLIP-based scores—often favored the incorrect view. Only our proposed metric, $\text{D}_{\text{PRISM}}$, consistently aligned with human preference, demonstrating superior sensitivity to semantic and geometric plausibility in unseen areas.}
    \label{fig:prism_success}
\end{figure*}

The study was conducted on 40 objects distinct from those in the \datasetname dataset. While \datasetname itself is constructed from 40 base objects, we selected a separate set of 40 objects exclusively for the user study to ensure no overlap with the evaluation benchmark. For each object, we selected two source–target pairs with large azimuthal rotations ($>90^\circ$), where plausibility differences are more salient and traditional metrics often fail. For each pair, we sampled four out of ten possible pairwise comparisons among five diverse NVS models: Zero123, Zero123-XL, SEVA, OpenLRM, and TRELLIS. Forty computer science students participated, each completing 15 tasks following a brief training phase with instructional examples. Tasks were shuffled across participants to ensure balanced coverage of the evaluation set. Representative success cases from the study are illustrated in \cref{fig:prism_success}, where $D_{\text{PRISM}}$ uniquely aligns with human preferences.

\section{Distributional Metrics Background}
\label{app:fd_mmd}
\textbf{Fréchet Distance (FD)} and \textbf{Maximum Mean Discrepancy (MMD)} are standard distributional metrics for evaluating generative models. Both operate on features extracted from pretrained backbones, without requiring paired ground truth.

\paragraph{Fréchet Distance (FD).}
The Fréchet distance between two distributions $P,Q$ over $\mathbb{R}^d$ is defined as the Wasserstein-2 distance:
\begin{equation}
    d^2_F(P,Q) := \inf_{\gamma \in \Gamma(P,Q)} \mathbb{E}_{(x,y)\sim \gamma} \|x-y\|^2 ,
\end{equation}
where $\Gamma(P,Q)$ is the set of couplings of $P,Q$.  
When both distributions are assumed Gaussian with means $\mu_r,\mu_g$ and covariances $\Sigma_r,\Sigma_g$, this simplifies to:
\begin{equation}
    d^2_F = \|\mu_r - \mu_g\|_2^2 + \mathrm{Tr}\!\bigl(\Sigma_r + \Sigma_g - 2(\Sigma_r \Sigma_g)^{1/2}\bigr).
\end{equation}
This Gaussian approximation underlies the widely used \emph{Fréchet Inception Distance (FID)}~\cite{heusel2017gans}, which uses InceptionV3 features.  
However, feature embeddings from deep networks are far from Gaussian, often multi-modal and heavy-tailed. As a result, FID is a biased estimator that can contradict human judgments, is sensitive to distortions and preprocessing choices, and requires large sample sizes for stability~\cite{jayasumana2024rethinking}. For this reason, we also report FD with DINOv2 features~\cite{oquab2023dinov2}, which provide more semantically aligned embeddings.

\paragraph{Maximum Mean Discrepancy (MMD).}
MMD measures the discrepancy between two distributions $P,Q$ via kernel embeddings in a reproducing kernel Hilbert space (RKHS). For a characteristic kernel $k$, the squared MMD is:
\begin{align}
    d^2_{\text{MMD}}(P,Q) &= \mathbb{E}_{x,x'\sim P}[k(x,x')] + \mathbb{E}_{y,y'\sim Q}[k(y,y')] \nonumber \\
    &\quad - 2\, \mathbb{E}_{x\sim P,\, y\sim Q}[k(x,y)].
\end{align}
Given samples $X=\{x_i\}_{i=1}^m$, $Y=\{y_j\}_{j=1}^n$, the \emph{unbiased} estimator is:
\begin{align}
    \widehat{d^2_{\text{MMD}}}(X,Y) &= \frac{1}{m(m-1)} \sum_{i\neq i'} k(x_i, x_{i'}) 
    \\ &+ \frac{1}{n(n-1)} \sum_{j\neq j'} k(y_j, y_{j'}) \nonumber \\
    &\quad - \frac{2}{mn} \sum_{i=1}^m \sum_{j=1}^n k(x_i, y_j).
\end{align}
Unlike FD, MMD makes no Gaussianity assumption, is unbiased, and can be computed efficiently on GPU. In practice, Gaussian RBF kernels are common, with bandwidth $\sigma$ set by the median heuristic. When applied to CLIP features, this is often referred to as \emph{CLIP-MMD}, and has been shown to correlate more strongly with human perception and to be more sample-efficient than FID~\cite{jayasumana2024rethinking}.
In our experiments, we adopt the unbiased estimator with an RBF kernel, using the median heuristic for bandwidth.


\end{document}